\documentclass[10pt,twocolumn,letterpaper]{article}

\usepackage[pagenumbers]{cvpr} 
\usepackage[accsupp]{axessibility}  

\usepackage{microtype}

\renewcommand{\paragraph}[1]{\vspace{.5em}\noindent\textbf{#1.}}

\usepackage{soul}
\setuldepth{foobar}

\usepackage{amsmath,amssymb,amsfonts}
\usepackage{algorithmic}
\usepackage{subcaption}
\usepackage{graphicx}
\usepackage{textcomp}
\usepackage{xcolor}
\usepackage{svg}
\usepackage{booktabs}
\usepackage{url}
\usepackage{balance}
\usepackage{multirow}
\usepackage{siunitx}
\svgsetup{ inkscapelatex=false,}

\definecolor{cvprblue}{rgb}{0.21,0.49,0.74}
\usepackage[pagebackref,breaklinks,colorlinks,allcolors=cvprblue]{hyperref}

\def\paperID{*****} 
\def\confName{CVPR}
\def\confYear{2026}

\title{InCaRPose: In-Cabin Relative Camera Pose Estimation Model and Dataset}

\author{
Felix Stillger\textsuperscript{1,2} \quad
Lukas Hahn\textsuperscript{2} \quad
Frederik Hasecke\textsuperscript{2} \quad
Tobias Meisen\textsuperscript{1} \\[0.35cm]
\textsuperscript{1}University of Wuppertal, \texttt{<lastname>@uni-wuppertal.de} \\
\textsuperscript{2}Aptiv, \texttt{<firstname>.<lastname>@aptiv.de}
}

\begin{document}
\maketitle

\begin{abstract}
Camera extrinsic calibration is a fundamental task in computer vision. However, precise relative pose estimation in constrained, highly distorted environments, such as in-cabin automotive monitoring (ICAM), remains challenging. We present InCaRPose, a Transformer-based architecture designed for robust relative pose prediction between image pairs, which can be used for camera extrinsic calibration. By leveraging frozen backbone features such as DINOv3 and a Transformer-based decoder, our model effectively captures the geometric relationship between a reference and a target view. Unlike traditional methods, our approach achieves absolute metric-scale translation within the physically plausible adjustment range of in-cabin camera mounts in a single inference step, which is critical for ICAM, where accurate real-world distances are required for safety-relevant perception. We specifically address the challenges of highly distorted fisheye cameras in automotive interiors by training exclusively on synthetic data. Our model is capable of generalization to real-world cabin environments without relying on the exact same camera intrinsics and additionally achieves competitive performance on the public 7-Scenes dataset. Despite having limited training data, InCaRPose maintains high precision in both rotation and translation, even with a ViT-Small backbone. This enables real-time performance for time-critical inference, such as driver monitoring in supervised autonomous driving. We release our real-world In‑Cabin‑Pose test dataset consisting of highly distorted vehicle‑interior images and our code at
\href{https://github.com/felixstillger/InCaRPose}{https://github.com/felixstillger/InCaRPose}.
\end{abstract}    
\section{Introduction}\label{sec:intro}
Accurately estimating the relative camera pose between two views is essential for applications such as SLAM, AR, and 3D reconstruction. Classical pipelines rely on geometric feature matching and epipolar geometry, which can degrade under occlusions or strong lens distortions. 
Convolutional and Transformer-based architectures represent a significant advance in deep learning, providing robust global representations.
However, state-of-the-art models that generalize well typically require large training sets for convergence and only work with specific camera intrinsics. Moreover, these models are unsuitable for edge deployment due to their prohibitive computational requirements and excessive parameter counts. 
\begin{figure}[t]
     \centering
         \includegraphics[width=0.475\textwidth]{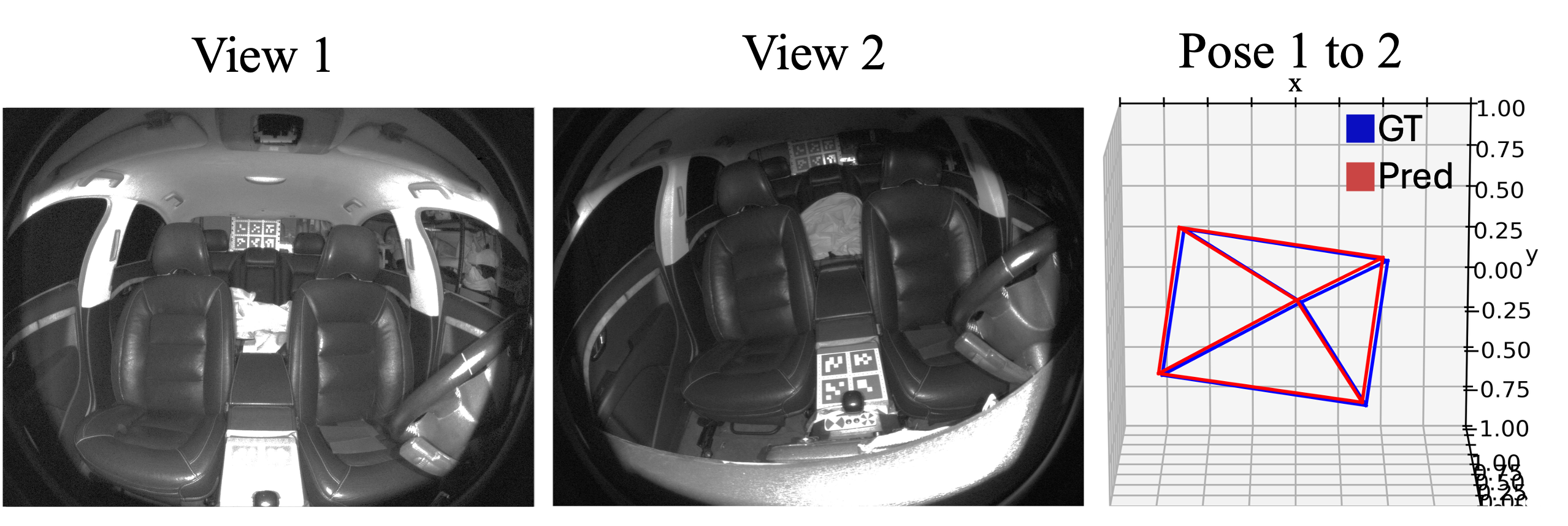}
         \caption{Our InCaRPose predicts the relative camera pose between a reference and a target view (shown as camera frustums). Trained exclusively on synthetic images, the model generalizes to real-world cabin environments and enables camera extrinsic calibration.}
        \label{fig:header}
\end{figure} 
The integration of in-cabin sensing cameras is accelerating, driven by EU GSR and Euro NCAP mandates, as well as rapid growth in driver and occupant monitoring~\cite{idtechex_incabin_2024,automation_dms_regulation_2023,rtinsights_incabin_2025}. These cameras are commonly mounted at multiple possible positions within the cabin, with the rear-view mirror being a preferred position due to its optimal view inside the cabin. When equipped with wide-angle or fisheye lenses, this configuration enables effective monitoring of both the driver and vehicle occupants in the front and back rows. However, rear-view mirrors are dynamic components that are manually or automatically adjusted by drivers, resulting in frequent changes to the camera’s extrinsic parameters. Maintaining consistent camera-to-vehicle calibration is a key challenge for reliable in-cabin perception. Interior cabin cameras typically operate in the near-infrared (NIR) spectrum rather than the more common visible RGB spectrum~\cite{piotrowski2022automotive}. Active NIR illumination (750–1400\,nm) ensures robust sensor visibility while remaining imperceptible to the human eye~\cite{naqvi2018gaze,rieder2021active} and therefore does not distract vehicle occupants. This mitigates the effects of cast shadows, surface texture variations, and ambient lighting changes, and enables night-time operation.
Accurate spatial awareness of in-cabin sensors is critical for driver monitoring, such as gaze estimation in autonomous driving to ensure drivers can return their hands to the wheel within required timeframes, as well as for other safety-relevant applications like occupant-position-aware airbag deployment. In the event of a collision, the system must infer occupant locations on the order of a few milliseconds to optimize restraint behavior because airbag control algorithms typically decide within 15–50\,ms (encompassing sensing-to-actuation pipeline) after detection~\cite{chidester2001recording,bortles2019performance,tsoi2013validation}.
We address camera extrinsic calibration in interior cabin environments and release our real-world in-cabin test dataset and code. Our model estimates the relative transformation between a potentially shifted cabin view and a calibrated reference view (Fig.~\ref{fig:header}). It combines frozen self-supervised ViT features with a Transformer decoder and a lightweight prediction head, enabling data-efficient training and real-time inference for time-critical automotive and industrial applications.
Training exclusively on a small synthetic cabin set is sufficient to recover large relative pose changes and generalize to real cabin imagery. We further evaluate the model's generalization capabilities on the public 7-Scenes and Cambridge Landmarks datasets, achieving competitive results without iterative solvers or dense 3D reconstructions.

\noindent In summary, our main contributions are:
\begin{itemize}
\item \textbf{A Novel In-Cabin-Pose Test Dataset.} We publish highly distorted, wide-FoV NIR fisheye imagery with metric ground truth as a real-world in-cabin test set.
\item \textbf{Vehicle-Agnostic Reference Formulation.} We reformulate the in-cabin pose estimation problem as reference-relative pose prediction to avoid vehicle-specific coordinate frames and individual retraining.
\item \textbf{Robust Fisheye Handling Without Undistortion.} InCaRPose processes distorted fisheye images end-to-end and predicts translation in absolute metric units.
\item \textbf{Synthetic-to-Real Capable Model.} Training on limited synthetic renderings yields strong transfer to a real cabin and competitive performance on 7-Scenes.
\end{itemize}

\section{Related Work}\label{sec:related}
\subsection{Absolute Pose Estimation}
Absolute pose methods localize a camera in a global scene by regressing the camera position in the world coordinate system from images. Typically, they use a 2D to 3D matching and Perspective-n-Point-algorithm (PnP) with RANSAC. End-to-end learning-based relocalizers encode the scene implicitly in neural network weights. This includes scene coordinate regression (SCR) methods~\cite{brachmann2017dsac, brachmann2018learning, brachmann2021visual} and absolute pose regression (APR), such as PoseNet~\cite{posenet2015posenet, kendall2017geometric}. Some of them use image retrieval to estimate pose approximation \cite{sattler2019understanding}. While effective, these often require scene-specific training or dense reconstructions. Minimal solvers and robust estimators continue to improve sampling efficiency for these tasks~\cite{ventura2024absolute}. 
More recently, several variants of absolute pose estimation aim to improve robustness and cross-scene generalization. DFNet~\cite{chen2022dfnet} improves absolute pose regression through direct feature matching, while Neural Refinement~\cite{chen2024neural} introduces a refinement stage that significantly improves pose accuracy. Map-relative pose regression~\cite{chen2024map} extends APR to settings where only sparse or lightweight maps are available. Furthermore, Yin et al.~\cite{yin2025towards} propose a Laplace-inspired distribution to better model rotation uncertainty in regression-based localization.
\subsection{Relative Pose Estimation}
Relative pose methods estimate the transformation between two views without explicit scene maps. Classical approaches rely on matching local features to recover the essential matrix~\cite{hartley2003multiple}, which is decomposed into relative rotation and a scaleless translation vector. This formulation has been enhanced by learned features~\cite{lowe2004distinctive,dusmanu2019d2} and by improved matchers such as SuperGlue~\cite{sarlin2020superglue} and LoFTR~\cite{sun2021loftr}. Robust estimators such as MAGSAC++~\cite{barath2020magsac} further improve reliability.

Deep relative pose regression (RPR) bypasses explicit correspondences to directly predict pose~\cite{melekhov2017relative, Laskar2017PoseNet}. Although some RPR methods use image retrieval~\cite{arandjelovic2016netvlad, humenberger2020robust} to look up posed database images, they often implicitly rely on prior SfM or SLAM reconstructions. To avoid reliance on fixed databases, Turkoglu et al.~\cite{turkoglu2021visual} proposed using graph neural networks to aggregate information across multiple views, utilizing relative pose supervision to improve re-localization accuracy. In contrast, map-free relocalization targets scenarios where only a single reference image is available~\cite{balntas2018relocnet, winkelbauer2021learning,arnold2022map}. 
ExReNet~\cite{winkelbauer2021learning} shows that extremely sparse reference sets (as few as four images) can cover indoor rooms and LENS~\cite{moreau2022lens} uses NeRF-based view synthesis to enhance relative localization. Lin et al.~\cite{lin2024learning} introduce neural volumetric pose features that fuse appearance and geometry for improved pairwise pose estimation. 
More challenging is the objective toward unseen scene generalization, where models are expected to localize in environments entirely absent from the training set. Idan et al.~\cite{idan2023learning} demonstrated that RPRs can effectively generalize to these unseen scenes, a concept that was further expanded to completely new landscapes in follow-up work~\cite{idan2024beyond}. Frameworks such as DirectionNet~\cite{directionnet2021} and, more recently, Reloc3r~\cite{reloc3r2025} further advance generalization.
\subsection{Pretrained Foundation Models}
A core challenge in pairwise pose estimation is the scale ambiguity of the translation. Multi-view pipelines can triangulate metric structure~\cite{zhang2006image}, while pairwise map-free settings must rely on additional priors such as depth~\cite{eigen2014depth}. Recent vision foundation models provide strong geometric cues via self-supervised pretraining on large-scale data. CroCo~\cite{croco2022} learns multiview consistency through cross-view completion. MAST3R and DUST3R~\cite{dust3r2023,mast3r2024} explicitly ground correspondence by predicting dense 3D structure from image pairs. Distillation-based ViTs such as DINOv2 and DINOv3~\cite{oquab2023dinov2,dinov3_2025} produce transferable dense features. Notably, DINOv2 variants can include learned register tokens, while DINOv3 adopts 2D rotary positional embeddings (RoPE)~\cite{su2021roformer} to improve flexibility across input resolutions. Positional encodings also explore explicit spatial tokens (SpatialFormer~\cite{xiao2024spatialformer}) and refined rotary embeddings for multimodal models (MHRoPE~\cite{huang2025mhrope}). DUNE~\cite{dune2025} introduces a universal encoder that further distills from the teacher models DINOv2, MAST3R and Multi-HMR~\cite{oquab2023dinov2,mast3r2024,baradel2024multi}. 
\section{Background}\label{sec:background}

\subsection{Problem Statement}
Modern vehicles increasingly deploy multiple interior cabin cameras at different locations. To fuse their outputs reliably, these sensors must operate in a shared reference frame, which requires accurate extrinsic calibration both per camera and relative to one another. Unlike fixed exterior sensors, some interior mounting points exhibit mechanical variability. A prominent example is a camera mounted on or near the rear-view mirror. This position offers a favorable cabin view, often with wide-angle or fisheye optics, but the mirror is adjustable and can change the camera extrinsics over time.
\begin{figure}[t]
     \centering
         \includegraphics[height=2.6cm]{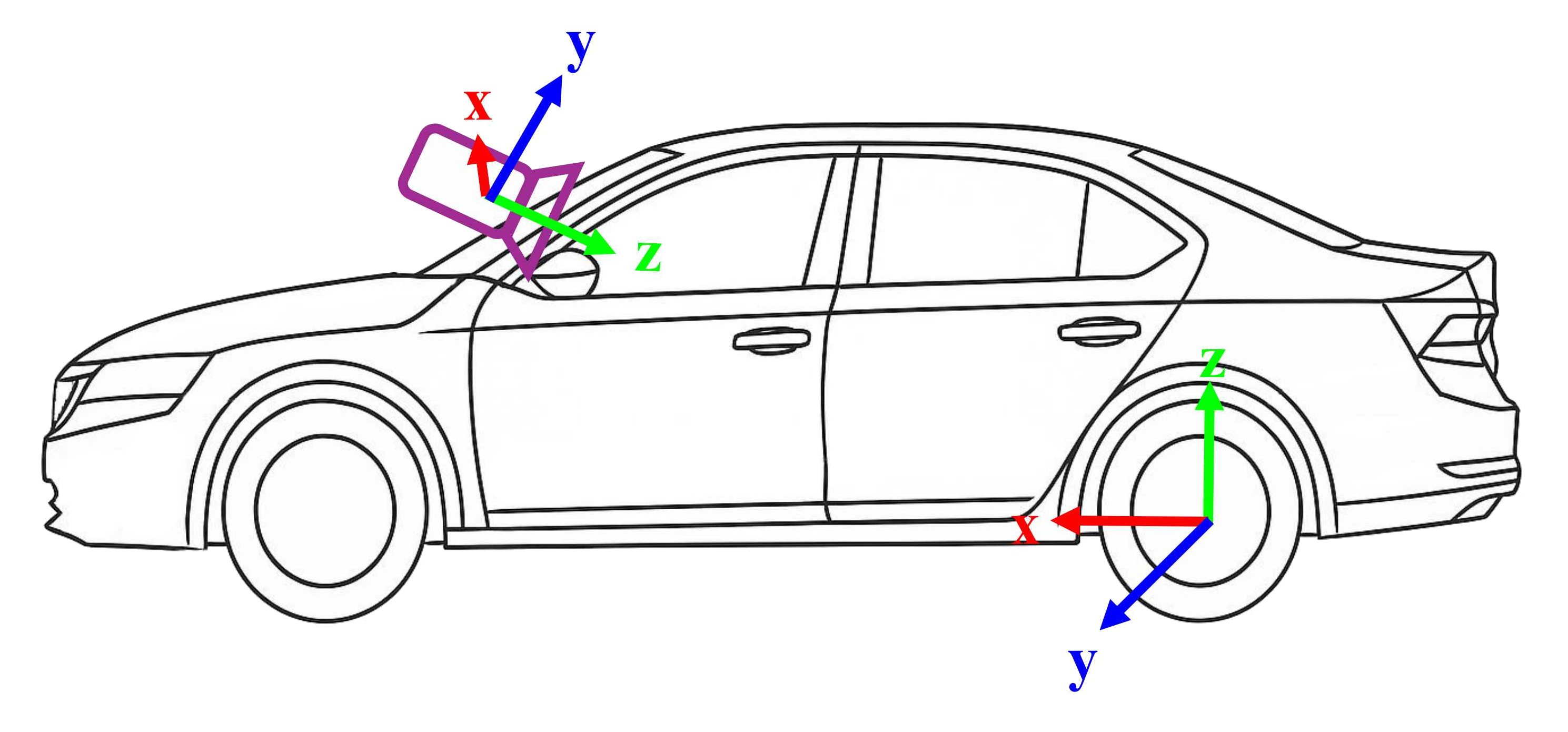}
         \caption{Camera coordinate system of the standard view compared to the vehicle's coordinate system.}
         \label{fig:cameras}
\end{figure}
As illustrated in Fig.~\ref{fig:cameras}, a standard calibration is commonly defined in the vehicle coordinate system (e.g., ISO~8855~\cite{ISO8855}). However, relying on such a global vehicle frame hinders generalization across vehicle models, since the geometric relationship between the cabin and the vehicle reference differs across platforms. To avoid vehicle-specific training, we reformulate the task as reference-relative pose estimation. Given a calibrated reference pose $\mathbf{T}_{v1}$ (e.g., a \emph{standard view} in the cabin) and a second view with pose $\mathbf{T}_{v2}$, we estimate the relative transformation $\mathbf{T}_{rel}$ such that
\begin{equation}
    \mathbf{T}_{v2} = \mathbf{T}_{v1} \cdot \mathbf{T}_{rel}.
    \label{eq:pose_decomposition}
\end{equation}
During training, $\mathbf{T}_{rel}$ is derived from ground-truth poses via $\mathbf{T}_{rel} = \mathbf{T}_{v1}^{-1}\mathbf{T}_{v2}$. At inference time, a single calibrated reference frame $\mathbf{T}_{v1}$ suffices, and the problem reduces to estimating the relative pose (rotation and translation) between a known calibrated state and a potentially shifted camera state from an image pair. This reference-relative formulation is vehicle-agnostic and enables deployment without retraining across different cabin configurations while supporting safety-relevant in-cabin perception.

\subsection{Data}\label{sec:data}
Our experiments rely on two data sources that consist of synthetic and real-world data. The synthetic data is used for training and validation, while the real-world data is exclusively for testing to approximate actual deployment conditions.
To collect the real-world dataset, the vehicle’s front windshield was removed, allowing free camera movement and preventing image disturbances caused by the confined cabin space. Controlling the camera from inside the vehicle would have influenced the data through operator-induced occlusions or motion artifacts. Because this process is invasive and difficult to repeat across multiple vehicles, collecting clean data in different car interiors is not straightforward. For this reason, our real-world test dataset is limited to a single vehicle cabin.
The real-world data includes experimental components, such as ArUco markers~\cite{garrido2014automatic}, which are not present in the synthetic data. A key advantage of the synthetic data is the availability of ideal ground truth without modifying the images, whereas ground truth generation for real-world data is inherently error-prone due to measurement inaccuracies and estimation uncertainty.
\begin{figure}[t]
    \centering
        \includegraphics[height=2.6cm]{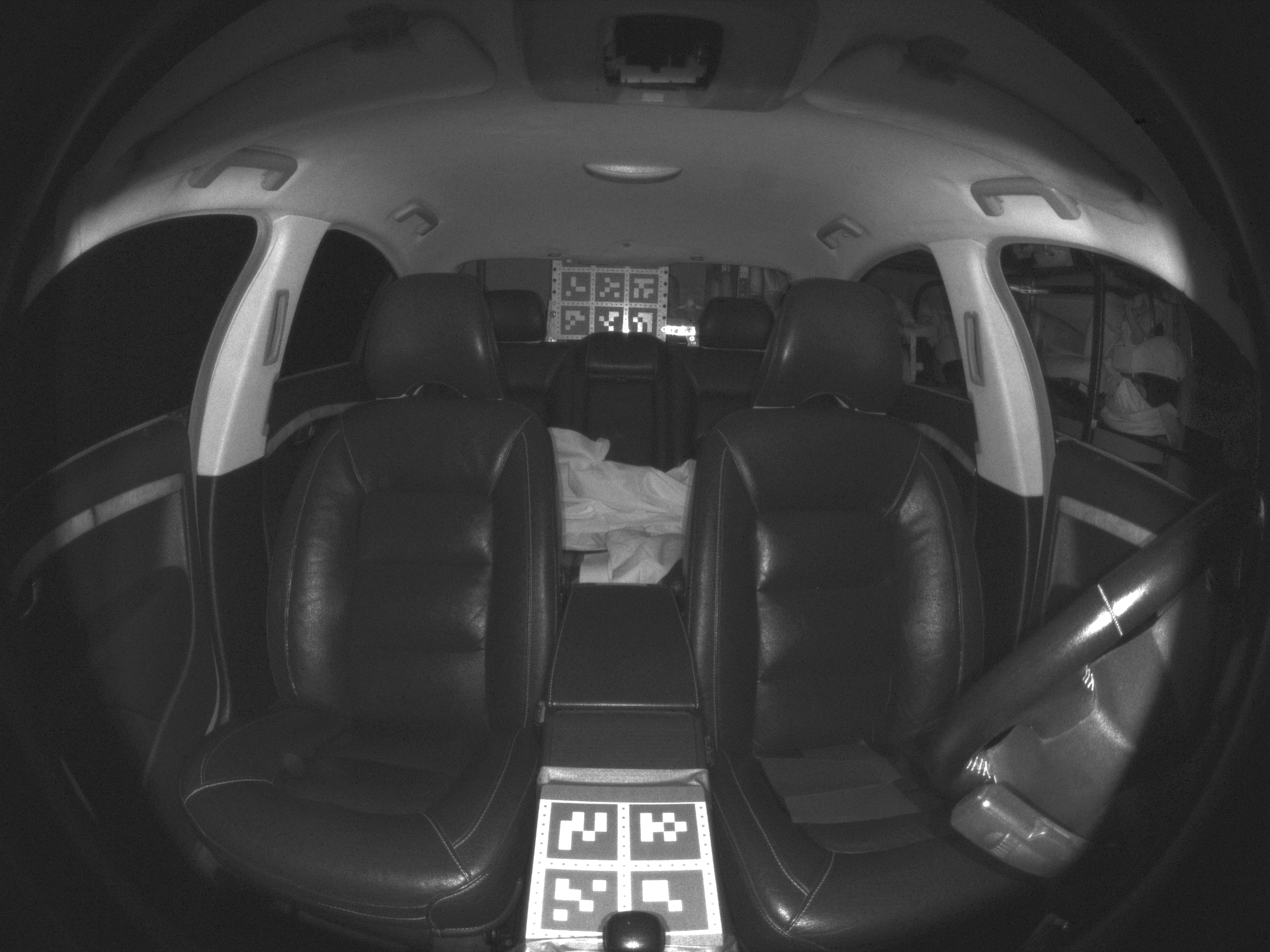}
        \includegraphics[height=2.6cm]{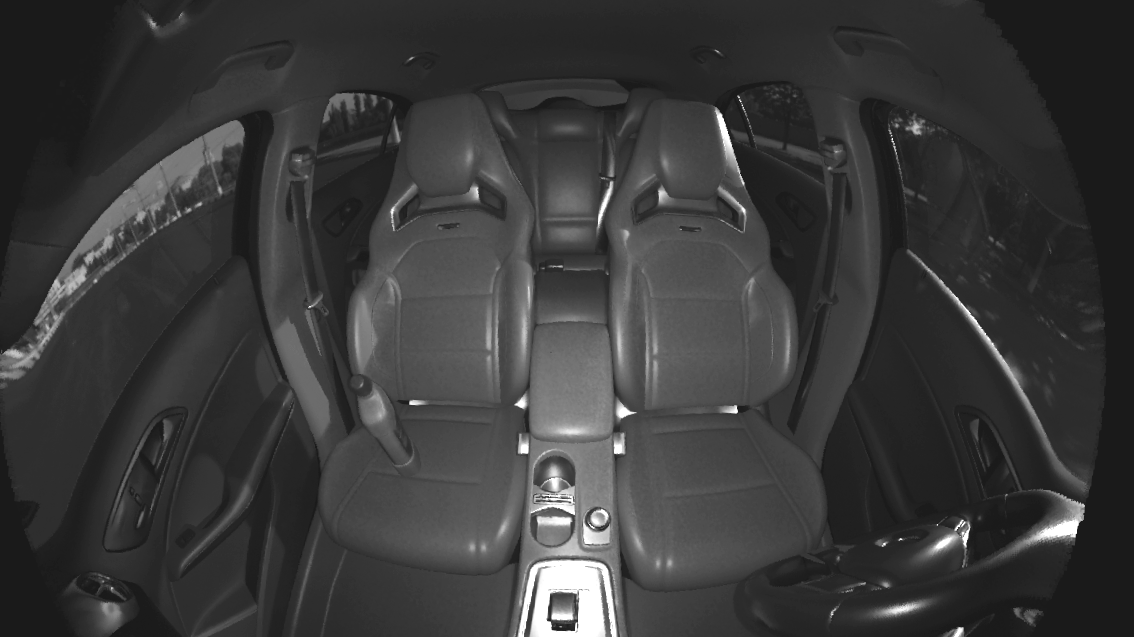}
    \caption{Standard view comparison: (left) real-world image and (right) synthetic image from the simulation environment.}
    \label{fig:comparison}
\end{figure}
Fig.~\ref{fig:comparison} shows a representative standard view for both real and synthetic data.

\paragraph{Synthetic Data} To cover a broad range of cabin geometries while retaining control over occlusions and distortions, we constructed synthetic scenes in Blender \cite{Blender2018}. We allocate eight vehicles for training and three for validation to avoid overlap between training and validation. We randomly place objects and occupants inside the cabin. Per scene, we uniformly sample rotations ($\pm80^\circ$ about $x$ and $y$, $\pm50^\circ$ about $z$) and translations ($\pm20$\,cm along each camera axis) around the per-vehicle standard view (Fig.~\ref{fig:comparison}). Although individual images are reused, we form unique image pairs, yielding approximately \mbox{$\sim$}5000 rotation-only, \mbox{$\sim$}1500 rotation+translation training pairs and \mbox{$\sim$}2000 validation pairs.

\paragraph{Real-World Data} Obtaining reliable ground truth in automotive interiors is challenging due to the confined space and the resulting low depth-to-baseline ratio. While Structure-from-Motion (SfM) frameworks like COLMAP~\cite{schoenberger2016sfm} are standard for large-scale scenes and can operate without visual cues such as ArUco markers, they suffer from scale ambiguity and geometric drift in environments. The lack of discriminative texture on monochrome surfaces often leads to ill-conditioned bundle adjustment leading to a non-optimal pose estimation in our particular use case.
To address this issue, we prioritize marker-based estimation. However, we provide an evaluation of both methods (see Supplementary Material~\ref{ap:groundtruth}). To obtain the camera pose, we detect ArUco markers of known size ($0.07\,\mathrm{m}$)~\cite{garrido2016generation} and estimate per-image marker-to-camera extrinsics using OpenCV's ArUco library~\cite{opencv_library}. In contrast to SfM, this yields metric scale and remains robust under the low-parallax conditions typical of confined interiors~\cite{opencv_library}. 
For each marker $m_k$ visible in both the reference image $r$ and the query image $j$, we obtain $\mathbf{T}^{(r,k)}_{c \leftarrow m_k}$ and $\mathbf{T}^{(j,k)}_{c \leftarrow m_k}$ and compute the relative camera pose as
\begin{equation}
\mathbf{T}^{(k)}_{c_j \leftarrow c_r}
=
\mathbf{T}^{(j,k)}_{c \leftarrow m_k}
\left(\mathbf{T}^{(r,k)}_{c \leftarrow m_k}\right)^{-1}.
\end{equation}

\paragraph{Real-World Test Dataset} To evaluate the robustness of our approach, we curated a real-world test dataset that systematically covers the camera's six degrees of freedom (6-DoF). This dataset is organized into specific sequences designed to isolate transformations and identify edge cases or model weaknesses. The sequences cover translations and rotations along each axis.
These axes are defined relative to the camera coordinate system. To ensure consistency across the dataset, we established a common reference frame based on a standard view within the cabin. This view was selected to maximize feature overlap across all subsequent transformations. All camera poses are computed as relative transformations from the origin of the standard view. We provide both ArUco- and COLMAP-generated ground truth. This results in 550 images and labels for our final released real-world test dataset called In-Cabin-Pose.

\section{Method}\label{sec:method}
\subsection{Architecture}
\begin{figure*}[t]
\centering
\includegraphics[width=\linewidth]{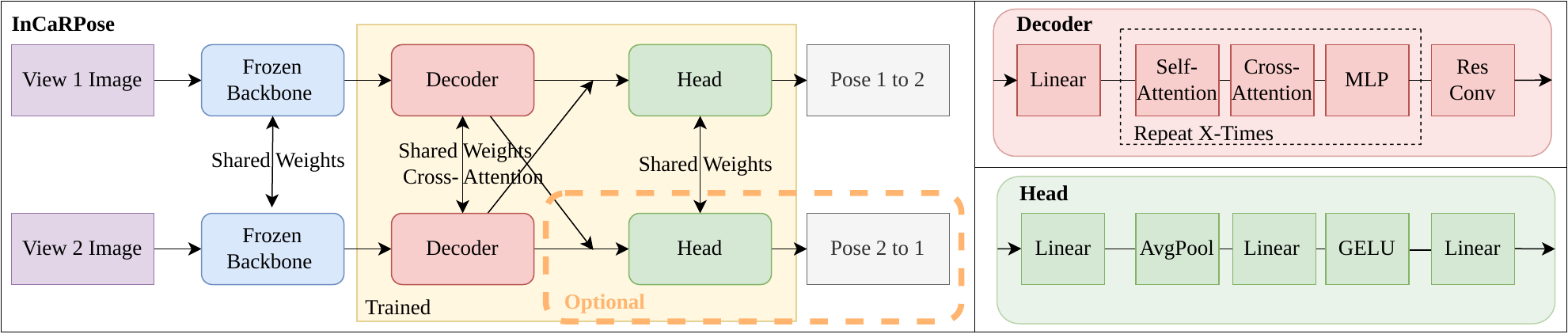}
\caption{InCaRPose's architecture overview. Two images are encoded by a frozen ViT backbone and fused by a cross-attention Transformer decoder. A prediction head outputs the relative camera pose between the views, optionally in both directions.}
\label{fig:arch}
\end{figure*}

We propose a two-view camera calibration network (see Fig. \ref{fig:arch}) that leverages a frozen vision Transformer backbone (e.g., DINOv3 \cite{dinov3_2025}) to extract per-patch latent features from a reference image and a second-view image. We utilize a frozen backbone because our task-specific dataset is limited in size. Through intensive pre-training, such models learn robust, cross-domain feature representations that can be effectively leveraged for a wide range of downstream tasks.
These features are further processed by a Transformer-based cross-attention decoder. This decoder is essential for refining the general-purpose feature relationships that are necessary for a relative pose estimation. To predict the final output, we utilize a compact multi-layer perceptron (MLP) head that regresses the representation-specific camera pose.
While the backbone remains frozen, only the decoder and prediction head are trained end-to-end. Maintaining a frozen backbone ensures that early training iterations do not degrade the fine-grained feature representations and preserves the model's ability to generalize across domains (e.g., from synthetic to real-world data). 

\paragraph{Details on Decoder and Prediction Head}
The model employs a Transformer decoder with variable depth to combine spatial features from the reference and second-view images. The tokens output by the backbone are linearly projected into decoder embeddings and processed by a stack of decoder blocks. To capture the spatial relationship between the two views, we designed each decoder block around self-attention to refine features and cross-attention to attend features to features across views. We add LayerNorm~\cite{vaswani2017attention} and residual paths to stabilize training. To incorporate spatial sensitivity without learned positional tokens, especially under limited training data, we introduce two-dimensional RoPE~\cite{su2021roformer} on queries and keys.
By default, the decoder uses twelve attention heads and an MLP expansion ratio of four, adopting a standard Transformer configuration that provides a balanced trade-off between computational efficiency and the capacity to model complex spatial dependencies. To reduce the feature-map dimensionality and connect the decoder with the prediction head, we use a residual convolutional bottleneck. This block performs channel-wise feature fusion to distill the high-dimensional Transformer embeddings, while the residual path maintains gradient flow and prevents information loss. These reduced features are then aggregated by global average pooling into a one-dimensional feature tensor. The final prediction head maps this low-dimensional tensor through a LayerNorm and a linear layer with a GELU activation function. We employ active dropout during training to improve generalization by preventing the model from overfitting to specific domain patterns. The final linear regression layer outputs a vector corresponding to the target output representation. To obtain a stronger supervisory signal and eliminate the need for image-ordering augmentation during training, we enable the model to predict the inverse relative pose. The final output is the concatenation of both outputs into a single tensor. This enforces geometric consistency by requiring the network to learn the bidirectional camera transformation relationship. During inference, double prediction can be disabled to maximize computational speed.

\subsection{Output Representations}

We also examine how the model adapts to different output styles and investigate various representations of rotations.

The common representation is $R \in \mathrm{SO}(3)$ which denotes a 3D rotation matrix, i.e., an orthogonal matrix with $\det(R)=1$ that preserves lengths and angles while maintaining the orientation of the coordinate system (i.e., it performs a proper rotation without reflection or flipping). Furthermore, $\mathbf{t} = [t_x, t_y, t_z]^\top \in \mathbb{R}^3$ denotes a translation vector. The combination of rotation and translation gives us a complete transformation that can be used to describe a relative pose.
We support five distinct parameterizations for the camera pose. In all cases, the translation is consistently appended as the final three components of the output vector $\mathbf{y}$. 
We consider multiple camera pose output representations, including rotation vectors and both intrinsic and extrinsic Euler angles with 3D rotation (6D total including translation), quaternions with 4D rotation (7D total), and rotation matrices with 9D rotation (12D total). Details and loss formulations are provided in Supplementary Material~\ref{ap:rot} and~\ref{ap:lossfunctions}.
To ensure valid rotations, we apply representation-specific post-processing. Depending on the target representation, we either normalize quaternions to unit quaternions with a magnitude of one, or we orthogonalize the rotation matrices via Singular Value Decomposition (SVD) with a det\,$=+1$ correction. These operations yield elements in $\mathrm{SO}(3)$, ensuring no scaling or shearing~\cite{levinson2020analysis,golub2013matrix,kuipers1999quaternions}.
\subsection{Labels and Preprocessing}\label{sec:training}
Datasets usually provide camera poses in dataset-specific global frames, e.g., the vehicle rear-axle frame (ISO~8855~\cite{ISO8855}) or an arbitrary world frame as in 7-Scenes~\cite{shotton2013scenecoords}. To make supervision independent of these conventions, we convert absolute poses into a canonical reference-relative target.
Given absolute poses for a reference view $T_{v1}$ and a second view $T_{v2}$, we define
\[
T_{\mathrm{rel}} \;=\; T_{v1}^{-1}T_{v2}, \qquad T_{v1}\equiv T_{\mathrm{ref}},
\]
which expresses the second view in the coordinate system of the reference view.
Because fisheye edges might contain critical geometric cues, we avoided center cropping. Instead, we rescale images and zero-pad them to maintain the full field of view even with varying input sizes (see Supplementary Fig.~\ref{fig:comparison_dataloaders}).
To increase the robustness against environmental changes, we use ColorJitter to manipulate the brightness, contrast, saturation, and hue of images. This prevents overfitting of the model on the clean synthetic data.
\section{Experiments}\label{sec:experiments}
We evaluate the synthetic-to-real generalization of \emph{InCaRPose} and report quantitative performance across our proprietary In-Cabin-Pose dataset, the public indoor 7-Scenes \cite{shotton2013scenecoords}, and public outdoor Cambridge Landmarks \cite{posenet2015posenet} dataset. To estimate a metric translation we use direct Euclidean distance error which minimizes the distance of predicted and ground truth translation. For rotational representation we utilize unit quaternion representation and its specific rotational loss (see Table \ref{tab:ablation_regular_gt}) since this yields the best performance. We train with a batch size of eight to enable parallel training on a single GPU and use the AdamW \cite{loshchilov2019decoupled} optimizer with a conservative learning rate of $1\times10^{-6}$ and weight decay of $1\times10^{-5}$ to prevent overfitting. The backbone is pretrained and frozen during training to preserve its robust, domain-invariant features \cite{oquab2023dinov2,dinov3_2025,dune2025} and to prevent their degradation by the noisy gradients of the randomly initialized components during the early training stages. Consequently, only the decoder and the head are subject to optimization.

\subsection{Keypoint-Based Detection}
We include a classical keypoint-based pipeline as a training-free geometric baseline and sanity check. Such methods are widely used for two-view pose estimation but typically recover translation only up to an unknown scale, which directly contrasts with our goal of metric translation. SIFT keypoints~\cite{lowe2004distinctive} are detected and matched using a Fast Library for Approximate Nearest Neighbors (FLANN) KD-tree~\cite{muja2009fast}, followed by filtering via Lowe’s ratio test with a threshold of $0.75$. Pairs with fewer than $20$ correspondences are discarded to ensure geometric reliability. For undistorted images, we estimate the Essential matrix~\cite{nister2004five} using the RANdom SAmple Consensus (RANSAC)~\cite{fischler1981random} algorithm to recover relative rotation ($R$) and translation ($t$). As SIFT is scale-ambiguous, translation is reported in terms of unit length only.

\subsection{Results on In-Cabin Data}
We first evaluate performance on our primary \textit{In-Cabin-Pose} dataset. Table~\ref{tab:model_comparison} reports results comparing InCaRPose with different frozen backbones (ViT-Small, -Base, -Large) against a classical SIFT matching pipeline and large-scale trained baselines (Reloc3r224 and Reloc3r512).
We note that Reloc3r predicts translation direction only and does not recover metric scale, and is trained on orders of magnitude more data than InCaRPose. When provided with undistorted images, Reloc3r achieves competitive directional translation performance, indicating that its strength lies in large-scale training under known camera intrinsics. In contrast, InCaRPose is designed to operate directly on highly distorted fisheye imagery and to recover metric-scale translation within the adjustment ranges encountered in in-cabin camera setups by a single forward pass, without requiring undistortion or scene-specific training.
The InCaRPoseLarge224 configuration achieves a median translation error of just $0.07\,\mathrm{m}$ and a median rotation error of 2.75$^\circ$. The low error rates indicate that the Transformer backbone effectively encodes global spatial relationships within the cabin. This robustness to real-world sensor noise, despite being trained exclusively on synthetic data, validates our approach to domain-invariant feature extraction.
SIFT-matching has competitive rotation performance but cannot determine absolute distance, which is a critical requirement for safety-critical automotive applications. Our InCaRPoseLarge224 model achieves the lowest overall rotation error, though it slightly underperforms Reloc3r in directional translation. The DUNE-Base encoder variant~\cite{dune2025}, with an input resolution of 504 due to its patch size, yields better results than DINOv3-Base, but further gains are constrained by the absence of a publicly released ViT-Large counterpart.
\begin{table}[t]
\centering
\caption{Model Performance Comparison: Rotation and Translation Errors (Trans. = translation error in m, Dir. = translation direction error in °) on our In-Cabin-Pose real-world test set.}
\label{tab:model_comparison}
\resizebox{\columnwidth}{!}{
\begin{tabular}{ll ccc}
\toprule
\textbf{Model} & \textbf{Error} & \textbf{Rot. ($^\circ$)} & \textbf{Trans. (m)} & \textbf{Dir. ($^\circ$)} \\
\midrule
\addlinespace
\multirow{2}{*}{InCaRPoseSmall224} & Mean & 6.11 & 0.11 & 42.55 \\
                                  & Median & 4.43 & 0.08 & 37.74 \\
\addlinespace

\multirow{2}{*}{InCaRPoseBase224} & Mean & 4.91 & 0.12 & 53.67 \\
                                 & Median & 3.55 & 0.09 & 42.45 \\
\addlinespace

\multirow{2}{*}{InCaRPoseLarge224} & Mean & 4.15 & \textbf{0.10} & 36.17 \\
                                  & Median & \textbf{2.75} & \textbf{0.07} & 23.46 \\
\addlinespace

\multirow{2}{*}{InCaRPoseDuneBase504} & Mean & \textbf{3.87} & 0.12 & 58.57 \\
                                     & Median & 3.05 & 0.09 & 39.71 \\
\midrule
\addlinespace

\multirow{2}{*}{Reloc3r224 \cite{reloc3r2025}} & Mean & 14.84 & -- & 83.43 \\
                                             & Median & 12.73 & -- & 76.79 \\
\addlinespace

\multirow{2}{*}{Reloc3r512 \cite{reloc3r2025}} & Mean & 13.65 & -- & 69.56 \\
                                             & Median & 11.48 & -- & 61.55 \\
\addlinespace

\multirow{2}{*}{Reloc3r224 \cite{reloc3r2025} (undistort)} & Mean & 4.74 & -- & \textbf{16.21} \\
                                                         & Median & 3.66 & -- & \textbf{11.63} \\
\addlinespace

\multirow{2}{*}{Reloc3r512 \cite{reloc3r2025} (undistort)} & Mean & 4.32 & -- & 17.74 \\
                                                         & Median & 3.23 & --& 13.05 \\
\addlinespace

\multirow{2}{*}{SIFT Matching \cite{lowe2004distinctive} (undistort)} & Mean & 7.15 & -- & 37.16 \\
                                                                    & Median & 4.83 & -- & 28.30 \\
\addlinespace
\multirow{2}{*}{Dust3r-Large \cite{dust3r2023} (undistort)} & Mean & 11.61 & -- & 19.60 \\
                                   & Median & 9.73 & -- & 12.13 \\

\addlinespace
\multirow{2}{*}{MAST3R-Large \cite{mast3r2024} (undistort)} & Mean & 14.24  & -- & 41.66 \\
                                   & Median & 11.27 & -- & 35.90 \\
\bottomrule
\end{tabular}
}
\end{table}
In addition to accuracy, the 224-resolution configurations are suitable for low-latency in-cabin perception under our evaluation setup. 
On a consumer-grade NVIDIA RTX\,4090 GPU (see Supplementary Table~\ref{tab:runtime_4090_full}), all InCaRPose variants run in real-time, consistently exceeding 45\,FPS, with the Small and Base backbones approaching 70\,FPS. 
Here, we use \emph{real-time} to denote batch-size-1 throughput sufficient to keep pace with a typical in-cabin camera stream (see Supplementary Table~\ref{tab:runtime_4090_full} for more details). InCaRPoseLarge224 achieves the best pose accuracy while remaining within real-time constraints despite its increased representational capacity. This illustrates the expected accuracy-latency trade-off.
At $\geq$67 \,FPS, the per frame \emph{model inference} latency is approximately 15 \,ms. 
However, 15–50 \,ms post-impact window for airbag deployment refers to the entire sensing-to-deployment pipeline, rather than an isolated perception module ~\cite{tsoi2013validation}. 
In practice, an airbag control system should therefore rely on the most recent \emph{pre-impact} (or immediately pre-trigger) occupant-state estimates maintained in a rolling buffer by the in-cabin perception stack, instead of waiting to execute a fresh pose inference after impact to satisfy tight response constraints.
For completeness, we evaluate inference on images without visible ArUco markers (see Supplementary Material chapter~\ref{sec:aruco-free}). Performance remains comparable, as expected, since markers are absent from the training data and the model does not depend on them. The model also remains robust to small distractors (a checkerboard is visible in one image only).
\subsection{Qualitative Results}
\begin{figure}[t]
     \centering
        \includegraphics[width=\linewidth]{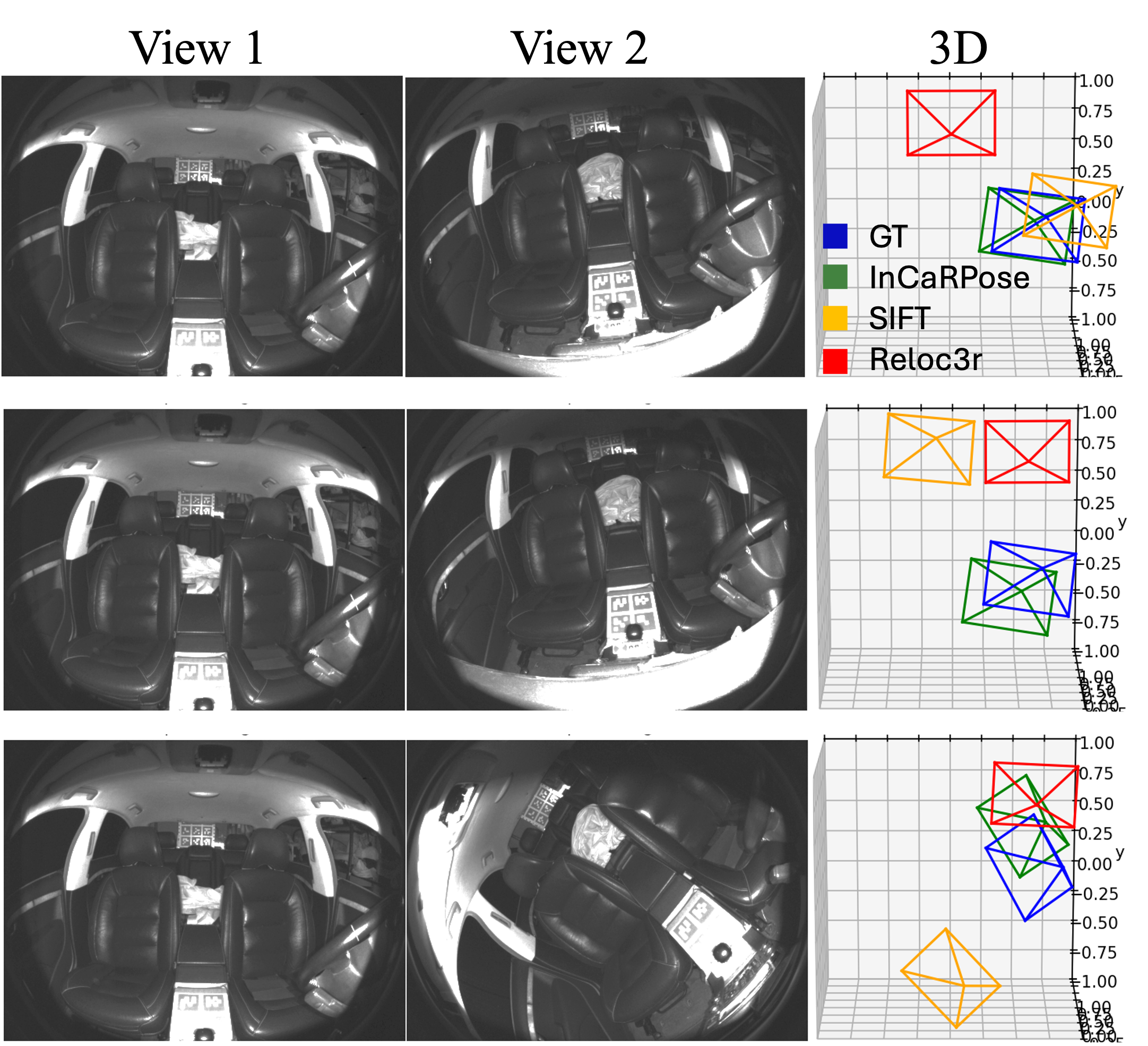}
        \caption{Qualitative results on real-world inference. All translation vectors are normalized to a common scale for visualization.}
        \label{fig:comparisoninf}
\end{figure}

Fig.~\ref{fig:comparisoninf} demonstrates InCaRPose on real-world examples. The predicted poses closely match the ground truth, indicating that the model captures spatial relationships even under challenging viewpoint changes and transfers effectively from synthetic training to real cabin scenes. We also visualize predictions from Reloc3r~\cite{reloc3r2025} and SIFT matching~\cite{lowe2004distinctive}, which show noticeably larger deviations in these cases. For visualization, all translation vectors are normalized to a common scale so that trajectories can be compared in a consistent coordinate frame. The last example highlights a remaining failure mode: although the model estimates rotation accurately, translation, particularly along the z-axis, remains challenging under extreme relative motion.
\subsection{Results on Public Data}
To evaluate how well our architecture generalizes to general indoor environments, we trained \textit{InCaRPose} on the 7-Scenes dataset using the training split defined by \cite{shotton2013scenecoords} in the original paper.
As shown in Table~\ref{tab:model_comparison_seven_scenes}, InCaRPose achieves substantially lower mean rotation error than Reloc3r and competitive or better median rotation error. Specifically, InCaRPoseLarge512 achieves a mean rotation error of 2.21$^\circ$, representing a 65\% improvement over the Reloc3r512 baseline (6.37$^\circ$). 
Furthermore, unlike Reloc3r, which is restricted to direction vectors, our model recovers the absolute metric translation. InCaRPose with ViT-Large backbones reaches the best median translation performance with 0.13m for the 224 input resolution. In contrast, the 512 version has slightly worse median translation performance but a lower rotational error. Our model also outperforms Reloc3r in directional translation error. The mean performance of our models is much closer to the median, suggesting that our model does not predict many drastic outliers. This also supports our use case. In contrast to the results on the In-Cabin-Pose dataset, the discrepancy between different backbones is not as large. This suggests that larger backbones can handle distortions more effectively, or that there is sufficient training data available to enable generalization of this dataset-specific problem, even for smaller backbones.
Additionally, we observe that increasing the input resolution only improves rotational performance, whereas translation performance does not improve with an increased input resolution.

\begin{table}[t]
\centering
\caption{Model performance: rotation and translation errors on 7-Scenes. Not available values are indicated with ``-''. Bold indicates best performance.}
\label{tab:model_comparison_seven_scenes}
\resizebox{\columnwidth}{!}{%
\begin{tabular}{ll ccc}
\toprule
\textbf{Model} & \textbf{Error} & \textbf{Rot. ($^\circ$)} & \textbf{Trans. (m)} & \textbf{Dir. ($^\circ$)} \\
\midrule
\multirow{2}{*}{Reloc3r224 \cite{reloc3r2025}\footnotemark[2]} 
  & Mean   & 7.96 & --       & 39.43 \\
  & Median & 2.45 & --       & 33.51 \\
\addlinespace

\multirow{2}{*}{RelPoseNet\cite{Laskar2017PoseNet}\footnotemark[1]} 
  & Mean   & --       & --       & -- \\
  & Median & 9.30     & 0.21     & -- \\
\multirow{2}{*}{Relformer\cite{idan2023learning}\footnotemark[1]} 
  & Mean   & --       & --       & -- \\
  & Median & 6.27     & 0.18     & -- \\
  
\addlinespace
\multirow{2}{*}{RelPoseGNN \cite{turkoglu2021visual}\footnotemark[1]} 
  & Mean   & --       & --       & -- \\
  & Median & 5.20     & 0.17     & -- \\
  
\addlinespace
\multirow{2}{*}{InCaRPoseSmall224} 
  & Mean   & 2.69 & 0.24 & 14.19 \\
  & Median & 2.39 & 0.14 & 9.32 \\
\addlinespace

\multirow{2}{*}{InCaRPoseBase224} 
  & Mean   & 2.56 & 0.24 & 14.36 \\
  & Median & 2.27 & 0.14 & 9.50 \\
\addlinespace

\multirow{2}{*}{InCaRPoseLarge224} 
  & Mean   & 2.55 & \textbf{0.22} & \textbf{13.48} \\
  & Median & 2.25 & \textbf{0.13} & \textbf{9.08} \\
\addlinespace

\midrule
\multirow{2}{*}{Reloc3r512 \cite{reloc3r2025}\footnotemark[2]} 
  & Mean   & 6.37 & --       & 38.80 \\
  & Median & 2.17 & --       & 33.14 \\
\addlinespace

\multirow{2}{*}{InCaRPoseLarge512} 
  & Mean   & \textbf{2.21} & \textbf{0.22} & 13.86 \\
  & Median & \textbf{1.91} & 0.15 & 10.35 \\

\bottomrule
\end{tabular}%
}
\end{table}
\footnotetext[1]{Values are taken from the authors' paper}
\footnotetext[2]{Trained on additional data}
To evaluate our model's performance on outdoor datasets, we present the results on the Cambridge Landmarks \cite{posenet2015posenet} dataset in Table \ref{tab:incarpose_cambridge_integrated}. We utilize the specific training and validation image pairs provided by RPNet \cite{en2018rpnet} for relative pose estimation, which are only available for the Kings College, Old Hospital, Shop Facade, and St Mary's Church subsets. The models were trained on all training data from all available scenes.
\begin{table}[t]
\centering
\small
\caption{Validation results on Cambridge Landmarks.}
\label{tab:incarpose_cambridge_integrated}
\resizebox{\columnwidth}{!}{%
\begin{tabular}{lll cccc}
\toprule
\textbf{Scope} & \textbf{Scene} & \textbf{Model} & \multicolumn{2}{c}{\textbf{Rot. Err ($^\circ$)}} & \multicolumn{2}{c}{\textbf{Trans. Err (m)}} \\
\cmidrule(lr){4-5} \cmidrule(lr){6-7}
 &  &  & Mean & Median & Mean & Median \\
\midrule

\multirow{4}{*}{Overall} 
 & \multirow{4}{*}{All} 
 & InCaRPoseSmall224 & 6.99 & 3.44 & 2.31 & 1.45 \\
 &  & InCaRPoseBase224  & 6.53 & 3.01 & 2.29 & 1.35 \\
 &  & InCaRPoseLarge224 & \textbf{6.49} & 2.66 & \textbf{2.19} & \textbf{1.21} \\
 &  & Reloc3r224 \cite{reloc3r2025}\footnotemark[2] & 9.95&\textbf{1.13}& -- & -- \\
\midrule

\multirow{20}{*}{Scene} 
 & \multirow{5}{*}{KingsCollege} 
 & InCaRPoseSmall224 & 2.74 & 2.67 & 1.20 & 1.05 \\
 &  & InCaRPoseBase224  & 2.53 & 2.29 & 1.15 & 1.09 \\
 &  & InCaRPoseLarge224 & 2.19 & 2.03 & \textbf{1.04} & \textbf{0.93} \\
 &  & Reloc3r224 \cite{reloc3r2025}\footnotemark[2] & \textbf{0.86} & \textbf{0.73} & -- & -- \\
 &  & RPNet \cite{en2018rpnet}\footnotemark[1] & -- & 5.40 & -- & 1.92 \\
\addlinespace

 & \multirow{5}{*}{OldHospital} 
 & InCaRPoseSmall224 & 3.62 & 2.96 & 1.55 & 1.45 \\
 &  & InCaRPoseBase224  & 3.16 & 2.54 & 1.40 & 1.32 \\
 &  & InCaRPoseLarge224 & 2.64 & 2.09 & \textbf{1.25} & \textbf{1.20} \\
 &  & Reloc3r224 \cite{reloc3r2025}\footnotemark[2] & \textbf{1.44} & \textbf{1.20} & -- & -- \\
 &  & RPNet \cite{en2018rpnet}\footnotemark[1] & -- & 5.40 & -- & 2.31 \\
\addlinespace

 & \multirow{5}{*}{ShopFacade} 
 & InCaRPoseSmall224 & 19.97 & 8.75 & 5.34 & 4.04 \\
 &  & InCaRPoseBase224  & \textbf{18.62} & \textbf{5.81} & \textbf{5.51} & \textbf{2.64} \\
 &  & InCaRPoseLarge224 & 20.31 & 9.31 & 5.52 & 3.76 \\
 &  & Reloc3r224 \cite{reloc3r2025}\footnotemark[2] & 42.70 & 11.78 & -- & -- \\
 &  & RPNet \cite{en2018rpnet}\footnotemark[1] & -- & 8.00 & -- & \textbf{1.46} \\
\addlinespace

 & \multirow{5}{*}{StMarysChurch} 
 & InCaRPoseSmall224 & 5.34 & 4.61 & 2.35 & \textbf{1.99} \\
 &  & InCaRPoseBase224  & 5.99 & 5.47 & 2.43 & 2.38 \\
 &  & InCaRPoseLarge224 & \textbf{4.73} & 4.66 & \textbf{2.32} & 2.22 \\
 &  & Reloc3r224 \cite{reloc3r2025}\footnotemark[2] & 11.69 & \textbf{2.27} & -- & -- \\
 &  & RPNet \cite{en2018rpnet}\footnotemark[1] & -- & 8.48 & -- & 2.65 \\

\bottomrule
\end{tabular}%
}
\end{table}
A performance disparity is observed when comparing these results to the 7-Scenes benchmark, with the Cambridge dataset proving more challenging. This degradation in precision can be attributed to the transition from indoor to outdoor environments. Indoor scenes typically offer higher feature density and consistency, whereas outdoor landmarks introduce significant scale variations and complex scenes.
The most notable loss of precision is on the Shop Facade subset. These findings suggest that the current architecture is better suited for environments characterized by high feature recurrence, which is more prevalent in structured indoor settings than in more variable outdoor scenes. This may stem from the RPNet \cite{en2018rpnet} split, which yields larger viewpoint and appearance gaps. We observe a similar degradation for Reloc3r \cite{reloc3r2025} on ShopFacade, which influences the overall mean performance drastically. Overall, Reloc3r attains lower rotation errors on several individual scenes, while InCaRPose is more competitive on ShopFacade and achieves stronger overall aggregated mean rotation.
\subsection{Ablations}

We have already demonstrated the effectiveness of different backbone sizes in Tables \ref{tab:model_comparison},\ref{tab:model_comparison_seven_scenes} and \ref{tab:incarpose_cambridge_integrated}, where InCaRPoseSmall has substantially fewer parameters than InCaRPoseLarge (see Supplementary Fig. \ref{fig:paramanalysis} for number of parameters against rotational error). For our In-Cabin-Pose data, the larger backbone improves quality, and the scope of the features helps learning with fewer examples. By contrast, the effect on the 7-Scenes and Cambridge Landmarks dataset is less significant. 
As shown in Table \ref{tab:model_comparison_seven_scenes}, increasing the resolution from 224 to 512 has a noticeable impact on the overall rotational performance, whereas the larger backbone in 224-resolution only has a marginal effect. The translation direction error and absolute metric error vary less than rotation across the backbones.

\paragraph{Output Representation}
We investigated different output mappings of the rotation (see Table~\ref{tab:ablation_regular_gt}).
\begin{table}[t]
\centering
\caption{Output rotation representations ablation on In-Cabin-Pose with InCaRPoseBase224.}
\label{tab:ablation_regular_gt}
\resizebox{0.98\columnwidth}{!}{%
\begin{tabular}{ll ccc}
\toprule
\textbf{Representation} & \textbf{Error} & \textbf{Rot. ($^\circ$)} & \textbf{Trans. (m)} & \textbf{Dir. ($^\circ$)} \\
\midrule
\multirow{2}{*}{Rotation Vector} & Mean & 6.35 & 0.14 & 81.24 \\
                               & Median & 4.80 & 0.11 & 70.62 \\ \addlinespace
\multirow{2}{*}{Euler Intrinsic} & Mean & 7.61 & 0.14 & 99.14 \\
                               & Median & 5.77 & 0.10 & 106.76 \\ \addlinespace
\multirow{2}{*}{Euler Extrinsic} & Mean & 6.39 & 0.14 & 108.73 \\
                               & Median & 4.67 & 0.11 & 110.92 \\  \addlinespace
\multirow{2}{*}{Quaternion} & Mean & 6.43 & 0.11 & 41.95 \\
                                     & Median & 4.89 & 0.07 & 36.23 \\ \addlinespace
\multirow{2}{*}{Rotation Matrix} & Mean & 5.83 & 0.16 & 71.97 \\
                               & Median & 5.03 & 0.13 & 73.67 \\ \addlinespace
\multirow{2}{*}{\textbf{Quaternion + Loss}} & \textbf{Mean} & \textbf{4.91} & \textbf{0.12} & \textbf{53.67} \\
                                     & \textbf{Median} & \textbf{3.55} & \textbf{0.09} & \textbf{42.45} \\ \addlinespace

\bottomrule
\end{tabular}%
}
\end{table}
Because prior work \cite{reloc3r2025,levinson2020analysis} suggested benefits of 9D parameterizations, we compare multiple rotational representations. Our goal was to identify the option that yields the best performance. Our ablation study was conducted on the ViT-Base size of DINOv3 \cite{dinov3_2025}. In our architecture, the configuration utilizing quaternions and the Quaternion-based Pose Loss achieved the lowest rotation error ($3.55^\circ$ median) compared to different rotation representations using the Universal Transformation Loss (see Supplementary Material chapter \ref{ap:rot}).

\paragraph{Backbone Type} 
Table~\ref{tab:backbone_ablation} compares different backbone architectures. 
\begin{table}[t]
\centering
\caption{Backbone ablation on In-Cabin-Pose with InCaRPoseBase224.}
\label{tab:backbone_ablation}
\resizebox{0.98\columnwidth}{!}{%
\begin{tabular}{ll ccc}
\toprule
\textbf{Backbone} & \textbf{Error} & \textbf{Rot. ($^\circ$)} & \textbf{Trans. (m)} & \textbf{Dir. ($^\circ$)} \\
\midrule
\multirow{2}{*}{Dune-Base}     & Mean & \textbf{4.52} & 0.13 & 55.35 \\
                              & Median & \textbf{3.00} & \textbf{0.09} & 53.40 \\
\addlinespace
\multirow{2}{*}{DUST3R-Large}  & Mean & 5.23 & 0.14 & 81.18 \\
                              & Median & 3.99 & 0.10 & 77.63 \\
\addlinespace
\multirow{2}{*}{DINOv2-Base}   & Mean & 5.98 & \textbf{0.12} & \textbf{45.78} \\
                              & Median & 3.95 & \textbf{0.09} & \textbf{40.43} \\
\addlinespace
\multirow{2}{*}{DINOv3-Base}         & Mean & 4.91 & \textbf{0.12} & 53.67 \\
                              & Median & 3.55 & \textbf{0.09} & 42.45 \\
\bottomrule
\end{tabular}%
}
\end{table}
As expected, we have noticed that the DINOv3 performs better than the DINOv2 as backbone. Like previously noticed, the Dune-Base encoder variant also shows strong performance in direct comparison, which is based on multiple teacher models, such as DINOv2, and outperforms our model with the DINOv3 backbone.

\section{Conclusion}\label{sec:conclusion}
We presented \emph{InCaRPose}, an effective relative pose estimator for highly distorted input images, primarily designed for automotive interiors but applicable beyond this setting. Relative pose estimation can be used to achieve absolute pose estimation with only a single calibrated frame to the vehicle coordinate system. Our architecture achieves competitive performance on the public 7‑Scenes and can generalize to an unseen real‑world car interior and an unseen camera, even when it is trained exclusively on synthetic data. It directly processes highly distorted fisheye images end-to-end, without any need for undistortion, preserving the original corner information.
A key outcome of our research is that the model is data‑efficient, achieving over 70\,FPS on a consumer GPU and successfully bridges the synthetic‑to‑real gap for real‑world inference, despite being trained on only 6,500 synthetic image pairs. This is achieved by leveraging powerful pre-trained backbones (such as the DINO-family) that remain completely frozen during training, enabling an efficient use of the features for other downstream tasks. Our model predicts translation in absolute metric units (meters) and performs competitively under challenging conditions.

Looking forward, we aim to enhance the model's robustness in partially occluded environments. While the current architecture has been trained on occlusion and obstacle data, future research will focus on developing a dedicated test set to quantify performance in complex, dynamic cabin conditions involving passengers and cargo.

\clearpage
{
    \small
    \bibliographystyle{ieeenat_fullname}
    \bibliography{main.bib}
}

\newpage
\clearpage
\section{Supplementary Material: InCaRPose}
\subsection{Supplementary Tables}

\begin{table}[h]
\centering
\caption{Detailed inference runtime measurements on a single NVIDIA RTX\,4090 GPU. We report average per-frame latency (ms), frames per second (FPS), and relative speedup with respect to the FP32 baseline at the corresponding backbone and resolution.}
\label{tab:runtime_4090_full}
\resizebox{\columnwidth}{!}{%
\setlength{\tabcolsep}{6pt}
\begin{tabular}{lll
                S[table-format=2.2]
                S[table-format=2.2]
                S[table-format=1.2]}
\toprule
\textbf{Backbone} & \textbf{Res.} & \textbf{Config} & {\textbf{Latency (ms)}} & {\textbf{FPS}} & {\textbf{Speedup}} \\
\midrule
Small & 224 & Baseline (FP32)               & 14.65 & 68.28 & 1.00 \\
      &     & FP16                          & 14.57 & 68.64 & 1.01 \\
      &     & \texttt{torch.compile}        & 14.00 & 71.41 & 1.05 \\
      &     & FP16 + \texttt{torch.compile} & 14.48 & 69.06 & 1.01 \\
\addlinespace
Small & 512 & Baseline (FP32)               & 24.75 & 40.40 & 1.00 \\
      &     & FP16                          & 16.49 & 60.65 & 1.50 \\
      &     & \texttt{torch.compile}        & 24.38 & 41.03 & 1.02 \\
      &     & FP16 + \texttt{torch.compile} & 14.76 & 67.75 & 1.68 \\
\addlinespace
Base  & 224 & Baseline (FP32)               & 14.91 & 67.06 & 1.00 \\
      &     & FP16                          & 14.93 & 66.97 & 1.00 \\
      &     & \texttt{torch.compile}        & 15.36 & 65.12 & 0.97 \\
      &     & FP16 + \texttt{torch.compile} & 14.46 & 69.16 & 1.03 \\
\addlinespace
Base  & 512 & Baseline (FP32)               & 31.18 & 32.07 & 1.00 \\
      &     & FP16                          & 17.62 & 56.75 & 1.77 \\
      &     & \texttt{torch.compile}        & 30.79 & 32.48 & 1.01 \\
      &     & FP16 + \texttt{torch.compile} & 17.62 & 56.74 & 1.77 \\
\addlinespace
Large & 224 & Baseline (FP32)               & 21.51 & 46.48 & 1.00 \\
      &     & FP16                          & 21.73 & 46.02 & 0.99 \\
      &     & \texttt{torch.compile}        & 17.04 & 58.69 & 1.26 \\
      &     & FP16 + \texttt{torch.compile} & 14.62 & 68.38 & 1.47 \\
\addlinespace
Large & 512 & Baseline (FP32)               & 62.49 & 16.00 & 1.00 \\
      &     & FP16                          & 30.44 & 32.85 & 2.05 \\
      &     & \texttt{torch.compile}        & 60.97 & 16.40 & 1.02 \\
      &     & FP16 + \texttt{torch.compile} & 24.70 & 40.49 & 2.53 \\
\bottomrule
\end{tabular}
}
\end{table}

\begin{table}[h]
\centering
\caption{Backbone ablation on COLMAP ground truth (distorted images and 224 resolution).}
\label{tab:backbone_ablation_colmap_distorted}
\resizebox{\columnwidth}{!}{%
\begin{tabular}{ll ccc}
\toprule
\textbf{Backbone} & \textbf{Metric} & \textbf{Rot. Err ($^\circ$)} & \textbf{Dir. Err ($^\circ$)} \\
\midrule
\multirow{2}{*}{Dune-Base}     & Mean & 6.26  & 52.11 \\
                              & Median & 3.88 & 50.90 \\
\addlinespace
\multirow{2}{*}{DUST3R-Large}  & Mean & 6.98 & 81.52 \\
                              & Median & 5.29 & 73.93 \\
\addlinespace
\multirow{2}{*}{DINOv2-Base}   & Mean & 8.11 & 45.60 \\
                              & Median & 5.30 & 39.58 \\
\addlinespace
\multirow{2}{*}{DINOv3-Base}         & Mean & 8.36 & 48.74 \\
                              & Median & 6.14 & 41.23 \\
\bottomrule
\end{tabular}%
}
\end{table}

\subsection{Rotation Representation}
\label{ap:rot}
We investigate different rotation representations and show how each can be mapped back to a uniform rotation matrix for the Universal Loss as described below:
\begin{enumerate}

    \item \textbf{Rotation Vector} ($\mathbb{R}^3$): 
    The rotation is represented by a compact axis-angle vector $\mathbf{\omega}$. The vector's direction specifies the rotation axis $\mathbf{u}$, and its magnitude represents the rotation angle $\theta = \|\mathbf{\omega}\|_2$ in radians. The mappingto a rotation matrix $R$ is given by Rodrigues’ \cite{rodrigues1840lois} formula:
    \begin{equation}
        R = I + \frac{\sin \theta}{\theta} [\mathbf{\omega}]_\times + \frac{1 - \cos \theta}{\theta^2} [\mathbf{\omega}]_\times^2
    \end{equation}
    where $[\mathbf{\omega}]_\times$ is the skew-symmetric matrix of $\mathbf{\omega}$. The final output is $\mathbf{y} = [\mathbf{\omega}^\top, \mathbf{t}^\top]^\top$.

    \item \textbf{Euler Angles: Intrinsic Rotation} ($\mathbb{R}^3$): 
    We support intrinsic rotations (moving axes) using the standard $ZYX$ convention. Given angles $(\alpha, \beta, \gamma)$, the final rotation matrix is computed by successive rotations around the transformed axes:
    \begin{equation}
        R_{\text{int}} = R_z(\alpha) R_{y'}(\beta) R_{x''}(\gamma)
    \end{equation}
    The final output is $\mathbf{y} = [\alpha, \beta, \gamma, t_x, t_y, t_z]^\top$.

    \item \textbf{Euler Angles: Extrinsic Rotation} ($\mathbb{R}^3$): 
    Extrinsic rotations are performed around the fixed, global axes $(X, Y, Z)$. For a sequence $(\gamma, \beta, \alpha)$, the resulting matrix is:
    \begin{equation}
        R_{\text{ext}} = R_z(\alpha) R_y(\beta) R_x(\gamma)
    \end{equation}
    The final output is $\mathbf{y} = [\gamma, \beta, \alpha, t_x, t_y, t_z]^\top$.

    \item \textbf{Quaternions} ($\mathbb{R}^4$): 
    The rotation is represented by a unit quaternion $\mathbf{q} = [w, x, y, z]^\top$, where $\|\mathbf{q}\|_2 = 1$. The mapping to $R$ is defined as:
    \begin{equation}
        R = \begin{bmatrix} 
        1 - 2(y^2 + z^2) & 2(xy - wz) & 2(xz + wy) \\ 
        2(xy + wz) & 1 - 2(x^2 + z^2) & 2(yz - wx) \\ 
        2(xz - wy) & 2(yz + wx) & 1 - 2(x^2 + y^2) 
        \end{bmatrix}
    \end{equation}
    The final output is $\mathbf{y} = [\mathbf{q}^\top, \mathbf{t}^\top]^\top$.

    \item \textbf{Rotation Matrix} ($\mathbb{R}^{9}$): 
    The rotation is represented directly by the flattened elements of $R \in \mathbb{R}^{3 \times 3}$. The matrix must satisfy the constraints of the Special Orthogonal group:
    \begin{equation}
        \text{SO}(3) = \{ R \in \mathbb{R}^{3 \times 3} : R^\top R = I, \det(R) = +1 \}
    \end{equation}
    The final output is the flattened nine elements of $R$ followed by $\mathbf{t}$, resulting in $\mathbf{y} = [r_{11}, r_{12}, \dots, r_{33}, t_x, t_y, t_z]^\top$. \\
\end{enumerate}

If the rotation is described as rotation matrix we can describe the full relative transformation as follows:
\[
T_{\mathrm{rel}} \;=\; T_{\mathrm{view1}}^{-1}\,T_{\mathrm{view2}}, \qquad T_{\mathrm{view1}}\equiv T_{\mathrm{ref}}
\]

\[
T_{\mathrm{view}}=\begin{pmatrix}R_{\mathrm{view}} & t_{\mathrm{view}}\\[4pt]0 & 1\end{pmatrix}, \qquad
T_{\mathrm{rel}}=\begin{pmatrix}R_{\mathrm{rel}} & t_{\mathrm{rel}}\\[4pt]0 & 1\end{pmatrix}
\]

\[
T_{\mathrm{view1}}^{-1}=\begin{pmatrix}R_{\mathrm{view1}}^{\top} & -R_{\mathrm{view1}}^{\top}t_{\mathrm{view1}}\\[4pt]0 & 1\end{pmatrix}
\]

\begin{align}
R_{\mathrm{rel}} &= R_{\mathrm{view1}}^{\top} R_{\mathrm{view2}},\\
t_{\mathrm{rel}} &= R_{\mathrm{view1}}^{\top}\bigl(t_{\mathrm{view2}} - t_{\mathrm{view1}}\bigr).
\end{align}

\subsection{Error Metrics and Loss Functions}
The introduction of various transformation representations necessitates a systematic investigation into the loss functions and error metrics tailored to each output format. We observed that the choice of representation significantly impacts optimization behavior and that final model performance is highly sensitive to the specific objective function employed.

\subsubsection{\textbf{Individual Error Metrics}}

\paragraph{Geodesic Distance:} Measures the minimum rotation angle required to align the estimated rotation matrix $\mathbf{R}_{\text{est}}$ with the ground truth $\mathbf{R}_{\text{gt}}$:
\begin{equation}
e_{\text{rot\_geo}} = \arccos\left( \frac{\text{Tr}(\mathbf{R}_{\text{est}}^\top \mathbf{R}_{\text{gt}}) - 1}{2} \right)
\end{equation}

\paragraph{Quaternion Error:} Minimizes the angular distance on the hypersphere between unit quaternions $\mathbf{q}_{\text{est}}$ and $\mathbf{q}_{\text{gt}}$, accounting for the double-cover property of $SO(3)$:
\begin{equation}
e_{\text{rot\_quat}} = 2 \arccos\left( \left| \mathbf{q}_{\text{est}}^\top \mathbf{q}_{\text{gt}} \right| \right)
\end{equation}

\paragraph{Euclidean Distance:} Measures the absolute metric distance between translation vectors in meters:
\begin{equation}
e_{\text{trans\_eucl}} = \left\| \mathbf{t}_{\text{est}} - \mathbf{t}_{\text{gt}} \right\|_2
\end{equation}

\paragraph{Translation Direction Error:} Measures the angular difference between predicted and ground truth translation vectors, providing a scale-invariant metric:
\begin{equation}
e_{\text{trans\_dir}} = \arccos\left( \frac{\mathbf{t}_{\text{est}}^\top \mathbf{t}_{\text{gt}}}{\|\mathbf{t}_{\text{est}}\| \cdot \|\mathbf{t}_{\text{gt}}\|} \right)
\end{equation}

\subsubsection{\textbf{Composite Loss Functions}}
\label{ap:lossfunctions}
During training, these metrics are combined into the following loss formulations:

\paragraph{Universal Transformation Loss} This loss handles full transformation matrices $\mathbf{T} \in SE(3)$ (rotation and translation). We decompose the matrices back into rotation and translation components and apply a weighted sum of geodesic and Euclidean errors:
\begin{equation}
\mathcal{L}_{\text{universal}} = \mathbb{E}\left[ e_{\text{rot\_geo}} \right] + \alpha \cdot \mathbb{E}\left[ e_{\text{trans\_eucl}} \right]
\end{equation}
where $\alpha$ is a weighting factor to balance the different units.

\paragraph{Reloc3r Loss} Inspired by \cite{reloc3r2025}, this loss is designed for the estimation of relative poses, where translation is predicted as a unit vector. It combines geodesic rotation error with translation direction error:
\begin{equation}
\mathcal{L}_{\text{Reloc3r}} = \mathbb{E}\left[ e_{\text{rot\_geo}} + \alpha \cdot e_{\text{trans\_dir}} \right]
\end{equation}

\paragraph{Mean Squared Error (MSE) Loss} A standard baseline applied to raw output vectors $\mathbf{p} \in \mathbb{R}^d$. This is utilized for different output representations, such as Euler angles ($d=6$) or Quaternions ($d=7$):
\begin{equation}
\mathcal{L}_{\text{MSE}} = \frac{1}{d} \left\| \mathbf{p}_{\text{est}} - \mathbf{p}_{\text{gt}} \right\|_2^2
\end{equation}

\paragraph{Quaternion-based Pose Loss:} This is our primary loss for the metric estimation. It utilizes the quaternion error for orientation and either Euclidean distance (for metric pose) or direction error (for scale-invariant pose) for translation:
\begin{equation}
\mathcal{L}_{\text{quat}} = \mathbb{E}\left[ e_{\text{rot\_quat}} \right] + \alpha \cdot \mathbb{E}\left[ e_{\text{trans}} \right]
\end{equation}
The choice of $e_{\text{trans}}$ ($e_{\text{trans\_eucl}}$ or $e_{\text{trans\_dir}}$) allows the model to trade off between absolute metric accuracy and directional consistency.

\subsection{Datasets}
\label{ap:datasets}

\begin{figure}[t]
     \centering
     \begin{subfigure}[t]{0.48\columnwidth}
         \centering
         \includegraphics[width=\textwidth]{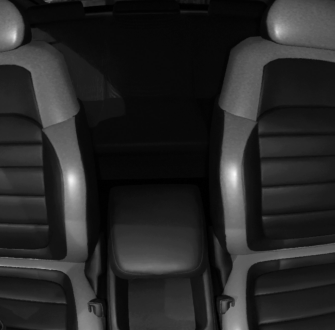}
         \caption{Center-crop: distorted}
         \label{fig:crop_distorted}
     \end{subfigure}
     \hfill
     \begin{subfigure}[t]{0.48\columnwidth}
         \centering
         \includegraphics[width=\textwidth]{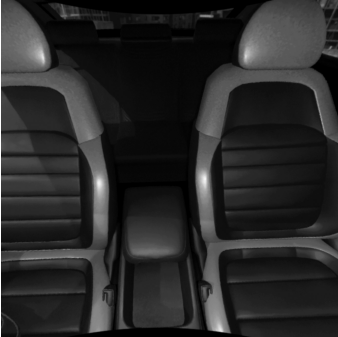}
         \caption{Center-crop: undistorted}
         \label{fig:crop_undistorted}
     \end{subfigure}

     \vspace{1em} 

     \begin{subfigure}[t]{0.48\columnwidth}
         \centering
         \includegraphics[width=\textwidth]{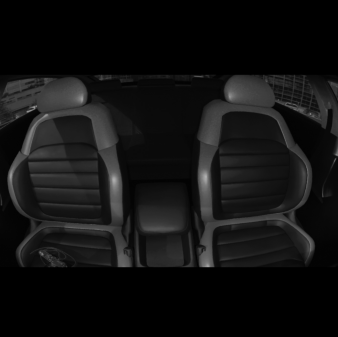}
         \caption{Zero-padded: distorted}
         \label{fig:pad_distorted}
     \end{subfigure}
     \hfill
     \begin{subfigure}[t]{0.48\columnwidth}
         \centering
         \includegraphics[width=\textwidth]{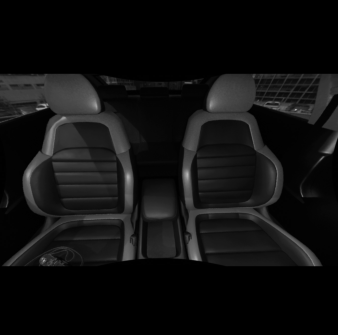}
         \caption{Zero-padded: undistorted}
         \label{fig:pad_undistorted}
     \end{subfigure}
     
     \caption{Comparison of preprocessing methods. (a) and (b): images are cropped to the center. (c) and (d): images are zero-padded to a square aspect ratio. This adjustment is necessary to handle varying input resolutions within the dataset while maintaining a consistent model input.}
     \label{fig:comparison_dataloaders}
\end{figure}

We show two different datasets in Fig. \ref{fig:comparison_dataloaders}. The center-crop dataset crops the image, resulting in data loss, while the zero-padded dataset does not cut the image. Instead, it pads the image to make it square. In both pipelines, images are converted to 8-bit RGB and, optionally, undistorted when intrinsics are available. The center-crop dataset resizes and center-crops to the target resolution, then applies standard ImageNet normalization. The zero-padded dataset rescales while preserving the aspect ratio, pads to a square canvas, and applies the same per-channel normalization.  The two datasets also demonstrate the trade-off between higher detail per resolution (center-crop dataset) and more border information (zero-padded dataset) while producing a predefined, fixed image size.

\subsection{Ground Truth Discussion}
\label{ap:groundtruth}
To assess the reliability of ground truth pose estimation in confined automotive interiors, we compare ArUco-based tracking with trajectories reconstructed using COLMAP.
Both translations are normalized to a uniform scale for a consistent comparison.
All sequences are expressed in a common local coordinate system defined by a single reference view within the vehicle cabin.
This reference view is selected to maximize feature overlap across all other views. This is important to maximize the likelihood that matching ArUco markers are visible across all scenes. All camera poses are represented as relative transformations with respect to this origin.

Table~\ref{tab:pose_errors} summarizes the observed discrepancies between COLMAP and ArUco-based estimates.
Rotation error is measured as the angular deviation between the COLMAP-estimated orientation and the ArUco-based ground truth.
Since COLMAP recovers translation only in unknown scale, translation error is evaluated exclusively in terms of direction, computed as the angular difference between the estimated and ground-truth translation vectors.
To avoid degenerate cases, translation direction error is reported only for frames where the relative Euclidean displacement of the ArUco ground truth exceeds 0.1\,m. We apply this threshold to ensure that only frames with meaningful translation contribute to this metric.

\begin{table}[t]
    \centering
    \caption{Rotation and translation direction errors. Translation direction error is only evaluated when the relative ArUco ground truth translation exceeds 0.1\,m.}
    \label{tab:pose_errors}
    \begin{tabular}{l c}
        \hline
        \textbf{Metric} & \textbf{Value} \\
        \hline
        Max rotation error & 19.86$^\circ$ \\
        Mean rotation error & 3.08$^\circ$ \\
        Median rotation error & 2.27$^\circ$ \\
        Max translation direction error & 55.83$^\circ$ \\
        Mean translation direction error & 5.75$^\circ$ \\
        Median translation direction error & 4.60$^\circ$ \\
        \hline
    \end{tabular}
\end{table}

\begin{figure}[t]
     \centering

         \includegraphics[width=\linewidth]{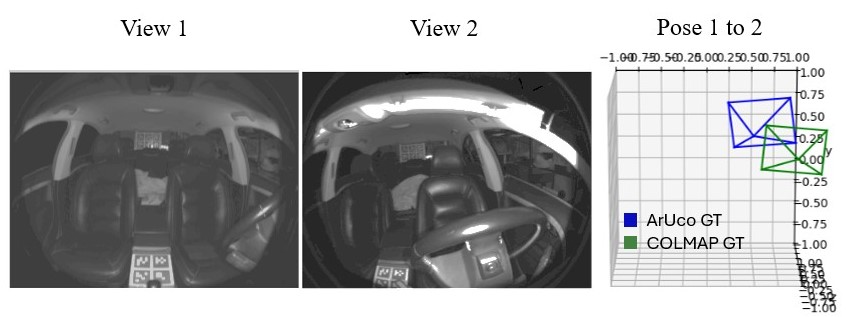}
         \caption{COLMAP fails to estimate translation along the $y$-axis. In the first view the camera is to the side of the steering wheel. In the second view the camera moved upwards (y) above the steering wheel. Translation is normalized.}
         \label{fig:comparison_aruco_colmap}

\end{figure}

\begin{figure*}[t]
\centering
\begin{subfigure}[b]{0.32\linewidth}
\includegraphics[width=\linewidth]{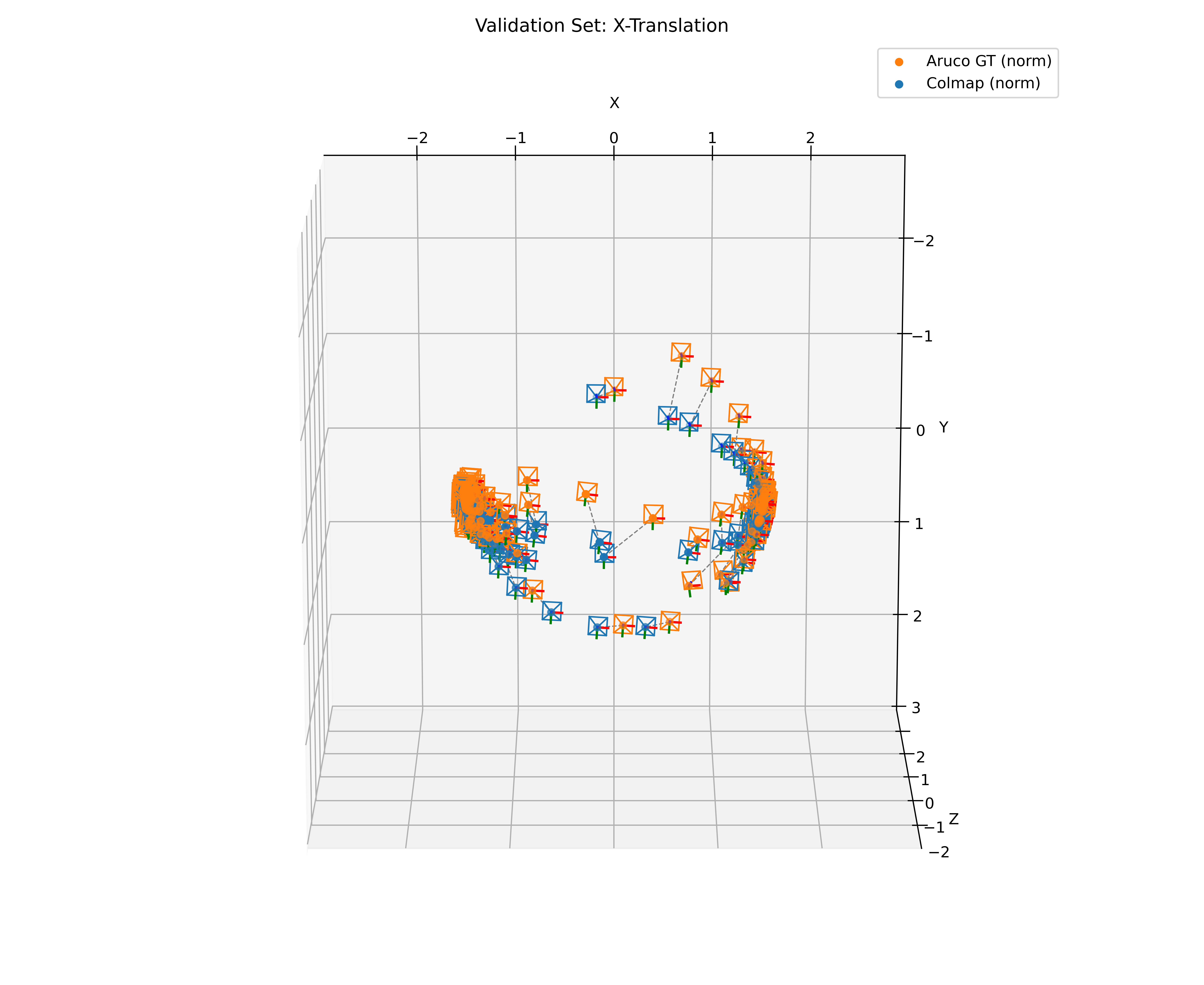}
\caption{X-Translation}
\end{subfigure}
\begin{subfigure}[b]{0.32\linewidth}
\includegraphics[width=\linewidth]{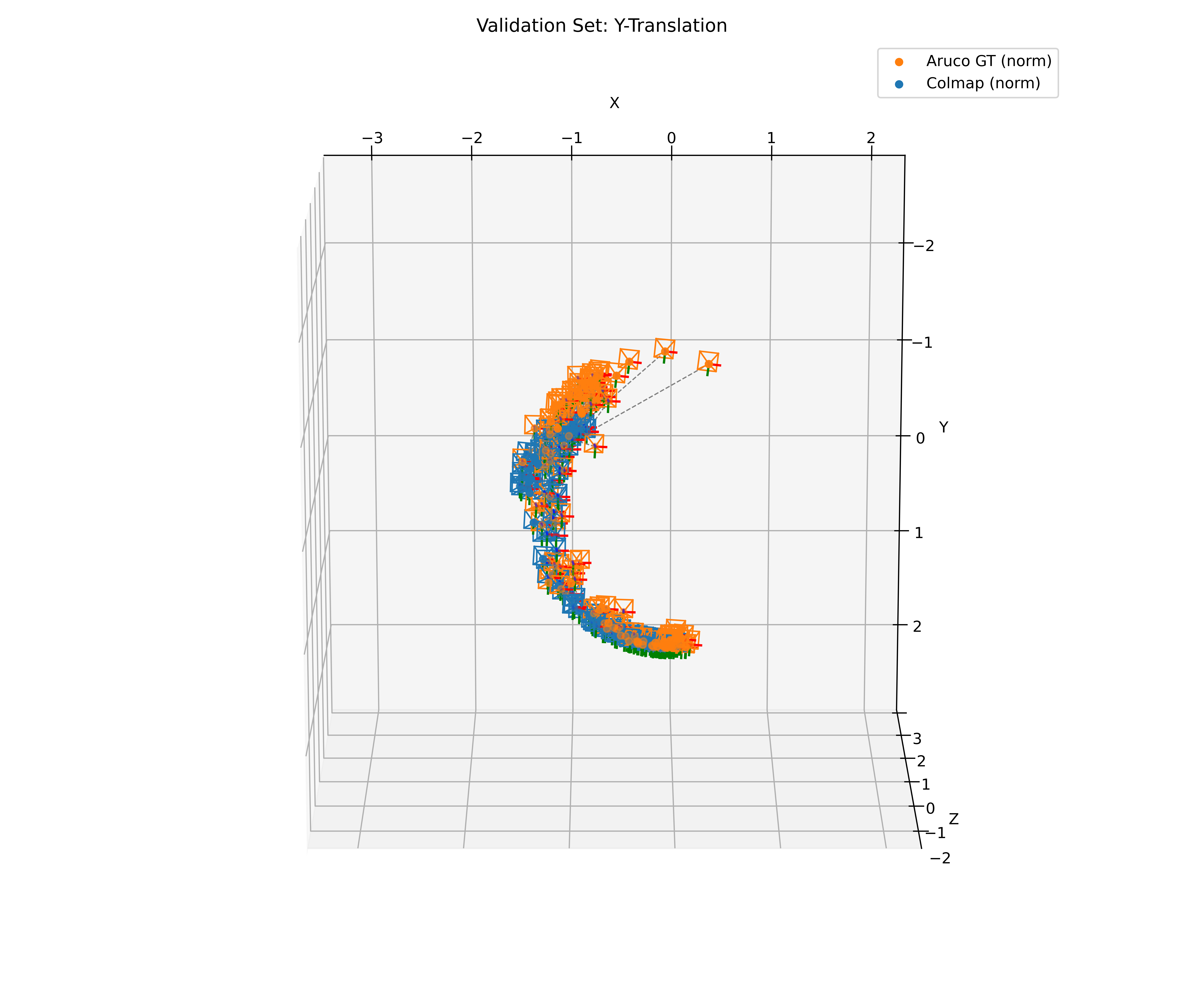}
\caption{Y-Translation}
\end{subfigure}
\begin{subfigure}[b]{0.32\linewidth}
\includegraphics[width=\linewidth]{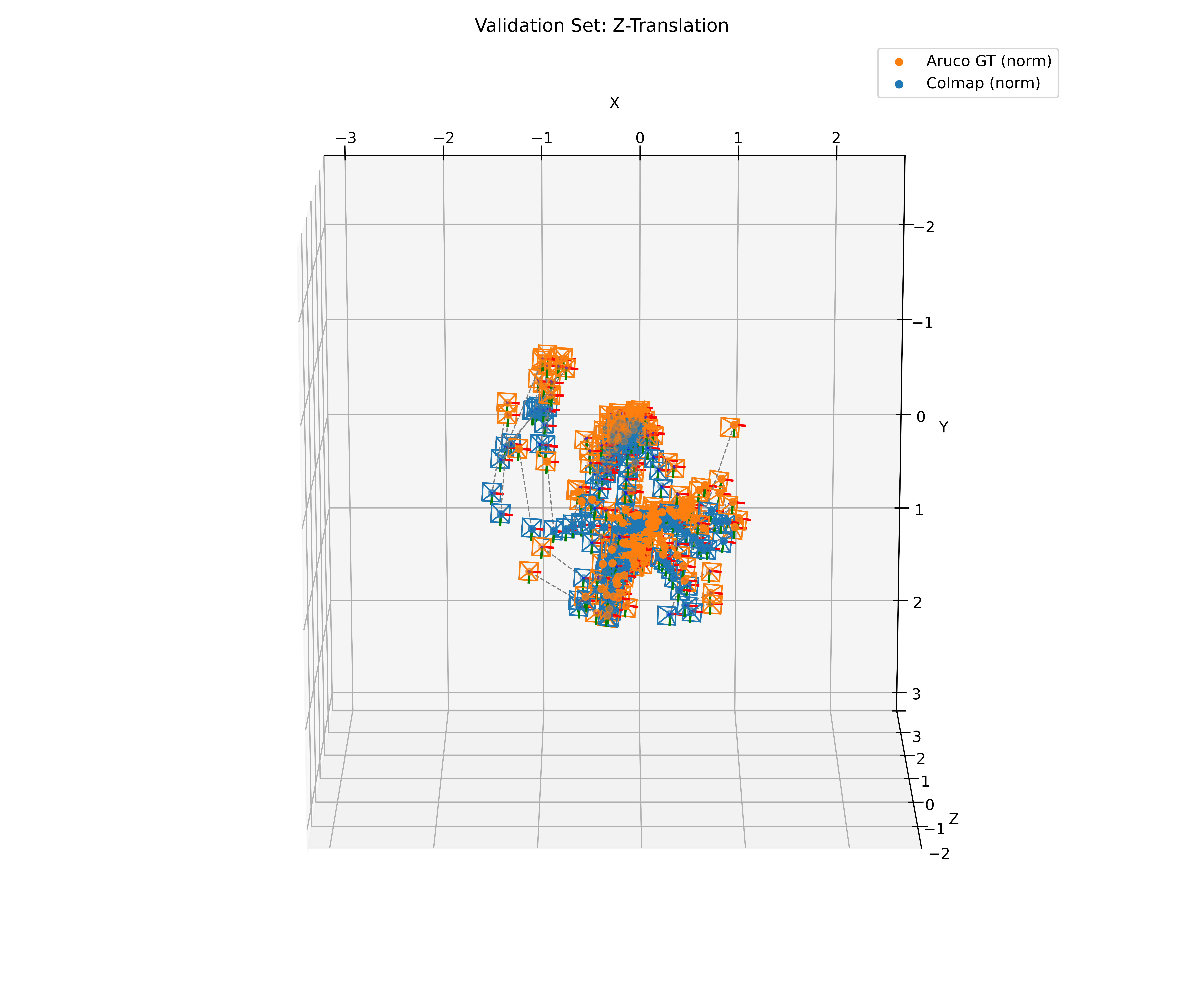}
\caption{Z-Translation}
\end{subfigure}

\vspace{4pt}

\begin{subfigure}[b]{0.32\linewidth}
\includegraphics[width=\linewidth]{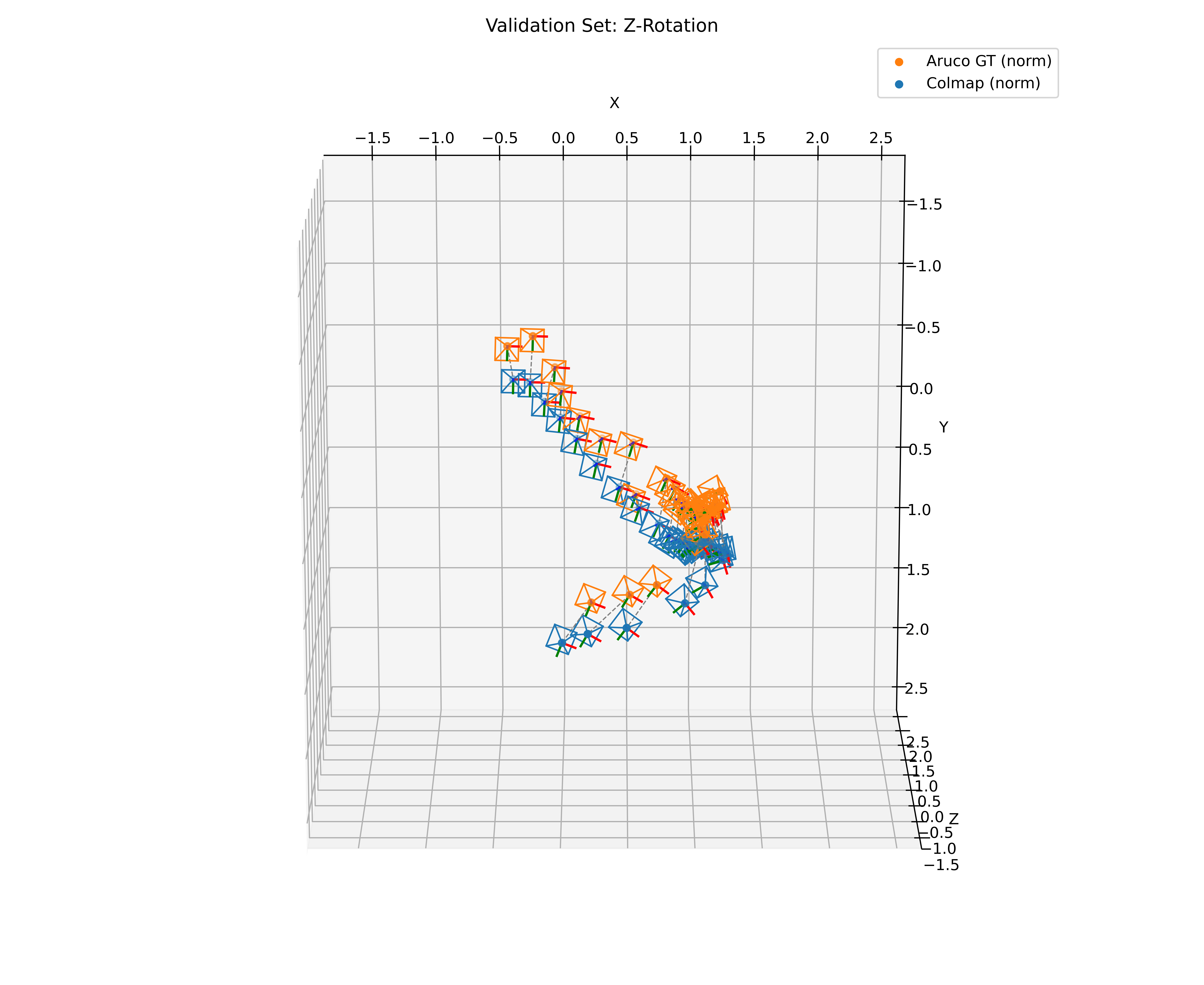}
\caption{Z-Rotation}
\end{subfigure}
\begin{subfigure}[b]{0.32\linewidth}
\includegraphics[width=\linewidth]{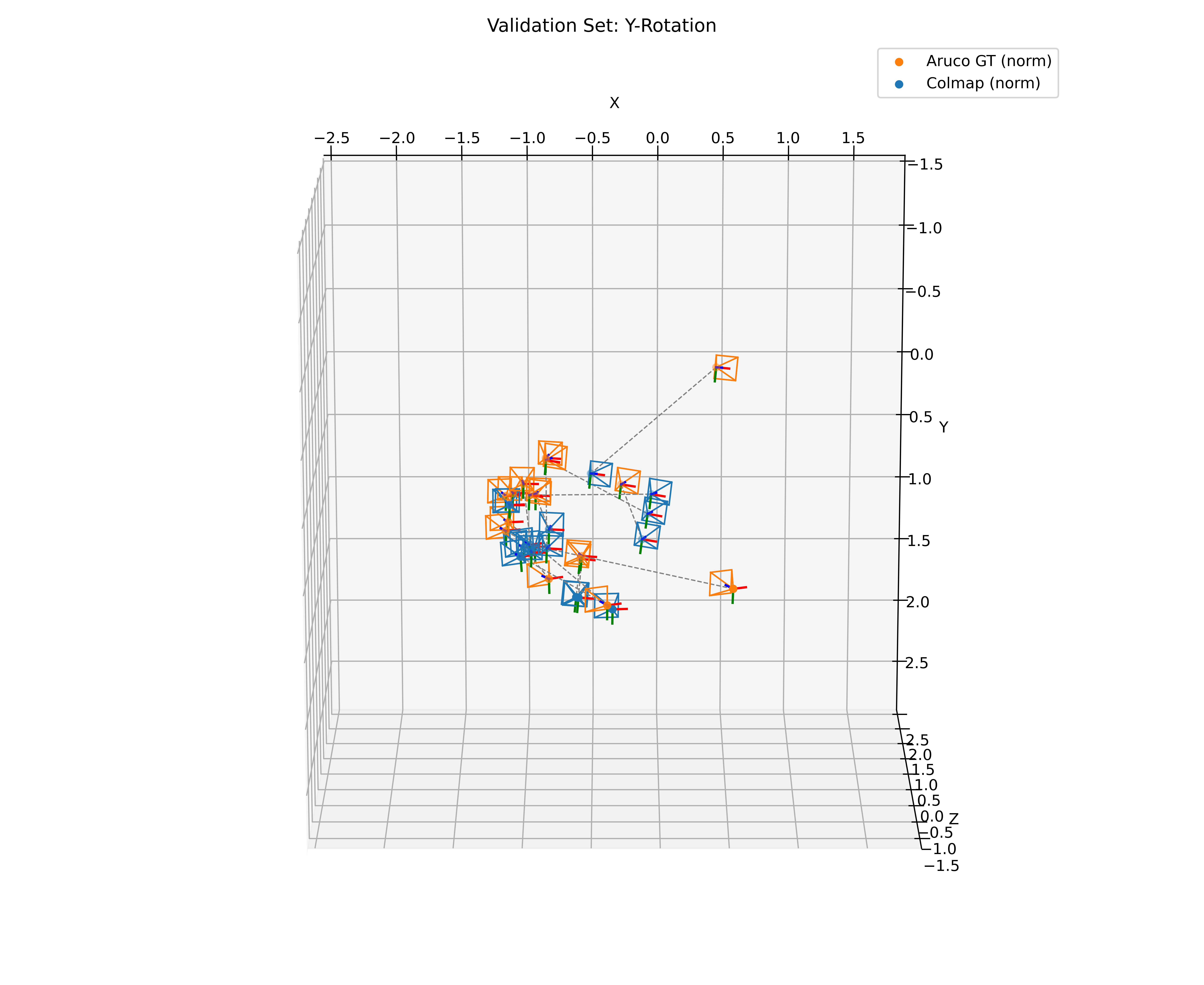}
\caption{Y-Rotation}
\end{subfigure}
\begin{subfigure}[b]{0.32\linewidth}
\includegraphics[width=\linewidth]{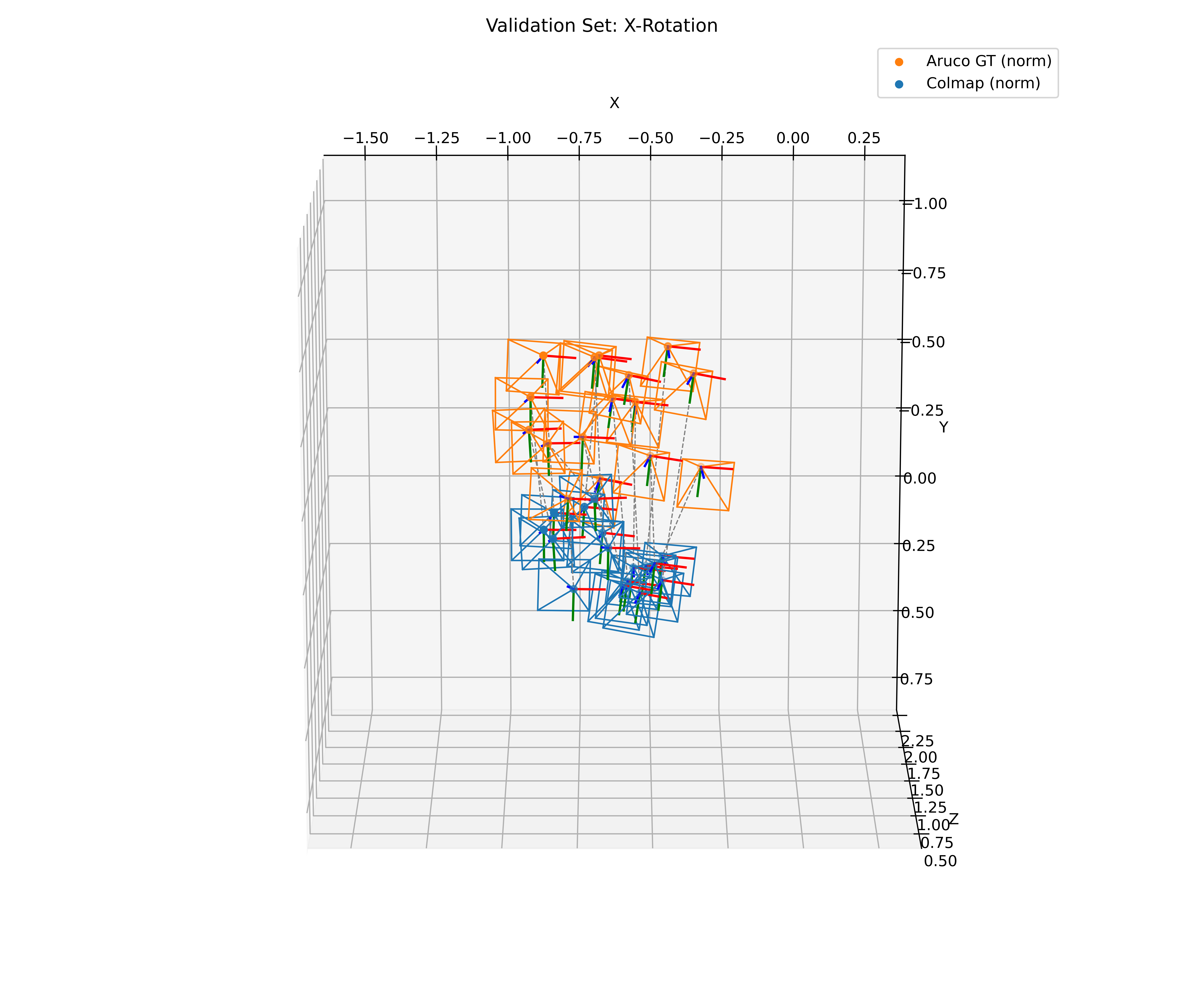}
\caption{X-Rotation}
\end{subfigure}

\caption{ArUco (orange) vs. COLMAP (blue) camera trajectories for intervals focused on specific transformations. Each sequence predominantly captures motion along the target axis, though residual degrees-of-freedom are also present.}
\label{fig:gt-colmap-trajectories}
\end{figure*}

As illustrated in Fig.~\ref{fig:comparison_aruco_colmap} and Fig.~\ref{fig:gt-colmap-trajectories}, discrepancies between the two methods arise in corner views.
In particular, COLMAP occasionally fails to recover vertical displacement (translation along the $y$-axis) or rotation around the $z$-axis, resulting in poses that are inconsistent with the physical camera placement.
Such failure cases are characteristic of confined interior environments, where limited baseline, weak texture, and reflective surfaces can lead to ill-conditioned structure-from-motion reconstructions.

Given that ArUco markers provide metric-scale translation and yield more physically plausible poses in these edge cases, we adopt the ArUco-based estimates as the primary ground truth for all quantitative evaluations in this work.
For completeness and reproducibility, we additionally release the corresponding COLMAP-based trajectories.

\begin{figure}[t]
     \centering
        \includegraphics[width=0.9\linewidth]{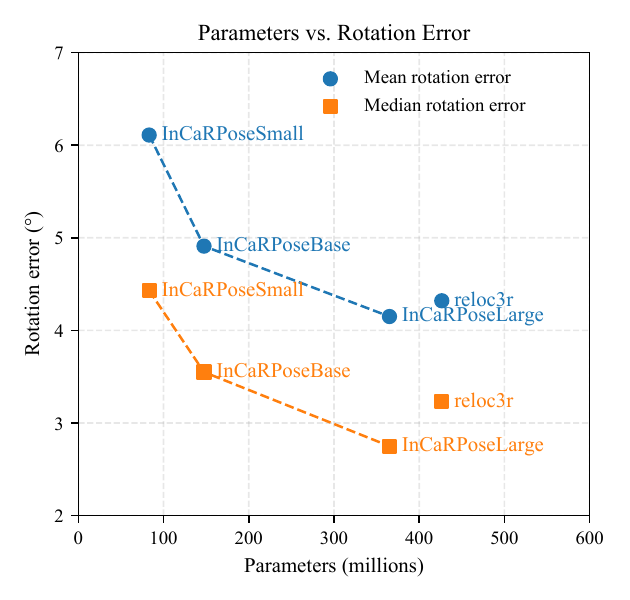}
         \caption{Rotation error in degrees versus the number of parameters. Evaluated on image resolution of 224 on the In-Cabin-Pose dataset.}
        \label{fig:paramanalysis}
\end{figure}

\subsection{ArUco-free Inference}
\label{sec:aruco-free}
\begin{figure*}[t]
\centering
\begin{subfigure}[b]{0.48\linewidth}
\includegraphics[width=\linewidth]{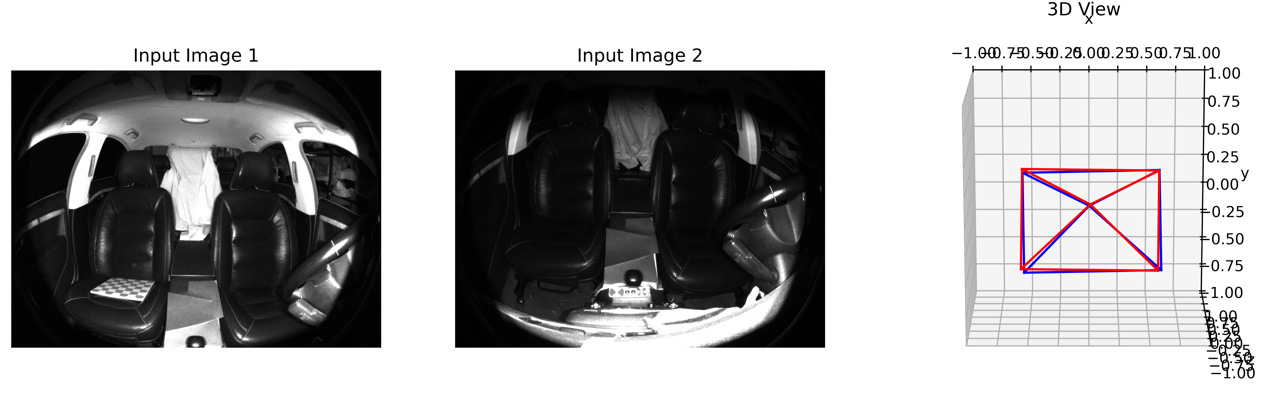}
\caption{}
\end{subfigure}
\begin{subfigure}[b]{0.48\linewidth}
\includegraphics[width=\linewidth]{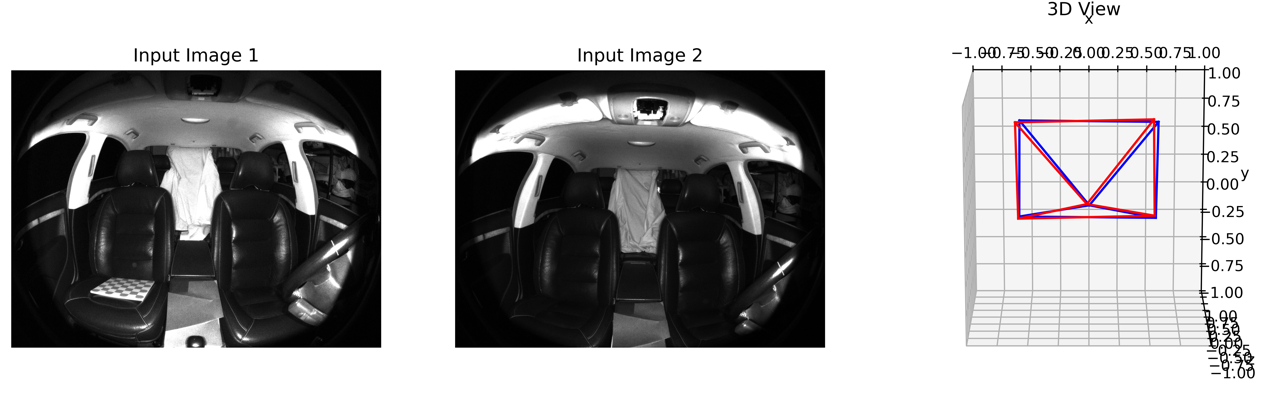}
\caption{Y}
\end{subfigure}

\vspace{4pt}

\begin{subfigure}[b]{0.48\linewidth}
\includegraphics[width=\linewidth]{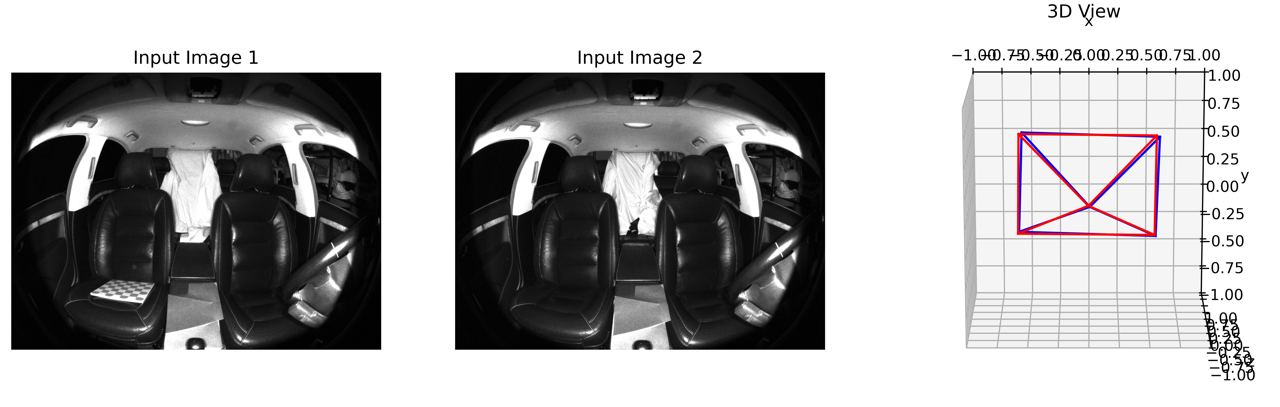}
\caption{Z}
\end{subfigure}
\begin{subfigure}[b]{0.48\linewidth}
\includegraphics[width=\linewidth]{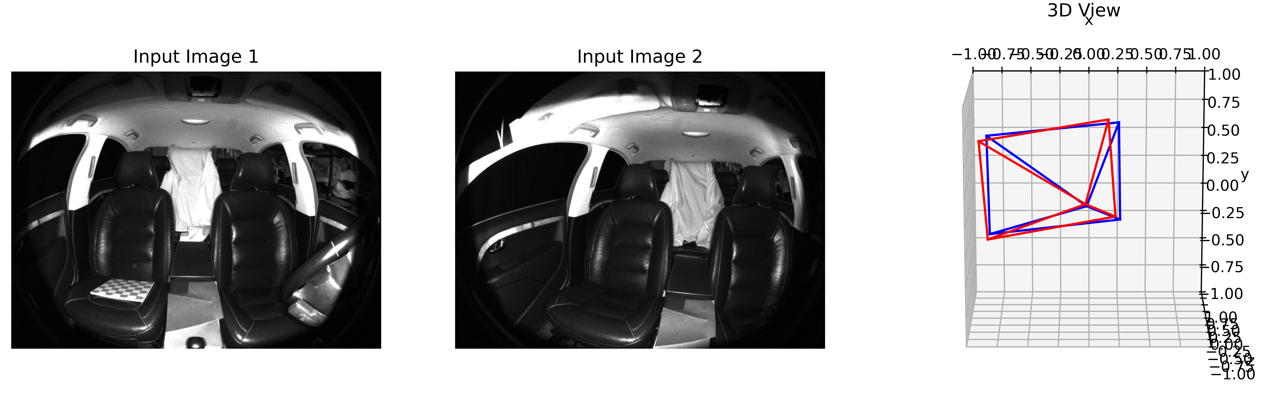}
\caption{Y}
\end{subfigure}

\caption{Inference on frames with physically occluded ArUco markers. We also changed the reference image, in which a different object (a high-contrast checkerboard on the left seat) is now visible in only one image to additionally challenge the model.}
\label{fig:aruco_occluded}
\end{figure*}

 We also provide several samples with occluded ArUco markers, shown in Fig. \ref{fig:aruco_occluded}. These examples demonstrate that the model does not rely on the presence of ArUco markers in the real‑world test data to make its predictions. For this data collection, we first captured frames in which the ArUco markers were fully visible. We then physically occluded the markers and recorded additional frames. Finally, we assigned the ground‑truth poses from the visible‑marker frames to the corresponding occluded‑marker frames.

\end{document}